\pdfoutput=1

\documentclass[11pt]{article}
\PassOptionsToPackage{table}{xcolor}
\usepackage[preprint]{acl}

\usepackage{times}
\usepackage{latexsym}

\usepackage[T1]{fontenc}

\usepackage[utf8]{inputenc}

\usepackage{microtype}

\usepackage{inconsolata}

\usepackage{graphicx}

\usepackage{microtype}
\usepackage{graphicx}
\usepackage{subfigure}
\usepackage{booktabs} 
\usepackage{hyperref}

\usepackage{algorithm}
\usepackage{algorithmic} 
\usepackage{longtable}
\usepackage{listings}
\usepackage{tcolorbox}
\usepackage{amsmath}
\usepackage{amssymb}
\usepackage{mathtools}
\usepackage{amsthm}
\usepackage{bm}
\usepackage[capitalize,noabbrev]{cleveref}
\theoremstyle{plain}

\theoremstyle{definition}

\theoremstyle{remark}

\usepackage[textsize=tiny]{todonotes}
\usepackage{multirow}
\definecolor{LightCyan}{rgb}{0.88,1,1}
\definecolor{YellowGreen}{rgb}{0.6, 0.8, 0.2}
\definecolor{GoldenRod}{rgb}{0.85, 0.65, 0.13}
\definecolor{WildStrawberry}{RGB}{120, 180, 230}

\title{A Training-Free Length Extrapolation Approach for LLMs:\\ Greedy Attention Logit Interpolation}

\author{
Yan Li$^1$, Tianyi Zhang$^2$, Zechuan Li$^3$, Soyeon Caren Han$^2$\thanks{Corresponding Author} \\
$^1$The University of Sydney, $^2$The University of Melbourne, $^3$Hunan University \\
$^1$\texttt{yali3816@uni.sydney.edu.au}, $^*$\texttt{caren.han@unimelb.edu.au}
}

\begin{document}
\maketitle
\begin{abstract}
Transformer-based Large Language Models (LLMs) struggle with inputs exceeding their training context window due to positional out-of-distribution (O.O.D.) issues that disrupt attention. Existing solutions, including fine-tuning and training-free methods, face challenges like inefficiency, redundant interpolation, logit outliers, or loss of local positional information. We propose Greedy Attention Logit Interpolation (GALI), a training-free method that improves length extrapolation by greedily reusing pretrained positional intervals and interpolating attention logit to eliminate outliers. GALI achieves stable and superior performance across a wide range of long-context tasks without requiring input-length-specific tuning. Our analysis further reveals that LLMs interpret positional intervals unevenly and that restricting interpolation to narrower ranges improves performance, even on short-context tasks. GALI represents a step toward more robust and generalizable long-text processing in LLMs. Our implementation of GALI, along with the experiments from our paper, is open-sourced at \url{https://github.com/adlnlp/Gali}.
\end{abstract}

\section{Introduction}
Transformer-based Large Language Models (LLMs) have become indispensable for a wide range of natural language processing tasks, yet their performance is fundamentally constrained by the training context window, i.e., the maximum input length used during training. When tasked with processing input text that exceeds this predefined limit, LLMs exhibit sharp performance degradation, with perplexity (PPL) increasing exponentially as input length grows \cite{ xiao2023efficient, han2024lm}. This limitation poses significant challenges for applications requiring robust long-text understanding, such as document summarization, legal text analysis, and conversational AI.

The core issue lies in the model’s inability to generalize beyond the positional distributions encountered during pretraining, leading to disruptions in attention score computations, a phenomenon known as positional out-of-distribution (O.O.D.) \cite{chen2023extending, Jin2024LLMML, xu2024base}. Addressing positional O.O.D. is critical for enhancing LLMs’ length extrapolation capabilities and enabling reliable long-text processing.

Existing approaches to mitigating positional O.O.D. can be classified into three categories: (1) Lambda-Shaped Attention Mechanisms, which stabilize PPL but compromise the ability to capture long-range dependencies across distant tokens \cite{xiao2023efficient, han2024lm, jiang2024minference, li2024quickllama}; (2) Fine-Tuning on long texts, which involves training on datasets with extended positional contexts using interpolation \cite{ding2024longrope, li2024extending, wu2024never} or extrapolation \cite{Zhu2023PoSEEC, Chen2023LongLoRAEF, ding2024longrope}. While effective, this approach is resource-intensive and still encounters cases where positional IDs exceed its fine-tuned context window; and (3) Training-free length extrapolation methods, which include Rotary Position Embedding (RoPE) frequency interpolation techniques (e.g., Neural Tangent Kernel (NTK), Dyn-NTK, YaRN) \cite{LocalLLaMA-2023-ntk, LocalLLaMA-2023-dyn-ntk, peng2023yarn} and inputs rearrangement strategies (e.g., SelfExtend, ChunkLlama) \cite{Jin2024LLMML, an2024training}. 

However, these training-free methods exhibit significant shortcomings: (a) They rely on a global scaling factor, leading to sensitivity and inconsistent performance across both long-context and short-context tasks. 
(b) methods like NTK, Dyn-NTK, and YaRN suffer from attention logit outliers due to their positional embedding interpolations; and (c) SelfExtend and ChunkLlama inherently disrupt local positional relationships, compromising model performance.



To overcome these limitations, we propose Greedy Attention Logit Interpolation (GALI), a novel training-free length extrapolation method. 

The innovations of GALI are twofold:
1) Greedy and Localised Interpolation: Instead of applying global scaling across all positions, GALI retains the pretrained positional IDs within the training context window, ensuring that performance on short-context inputs remains uncompromised. For tokens beyond the training context window, GALI performs interpolation at a fine-grained, token- or chunk-specific level. This greedy, localised approach eliminates redundant extrapolation, enabling stable handling of long-context tasks.
2) Logit-Level Interpolation with Positional Noise: Unlike prior work that manipulates positional embeddings, GALI operates on attention logit. It interpolates logit between valid positional pairs and injects Gaussian noise scaled to their positional interval. This design captures the oscillatory characteristics of RoPE while preventing numerical instability, resulting in robust length extrapolation.

With this, GALI explicitly addresses the shortcomings of existing methods by: (a) providing stable and superior performance on long-context tasks without compromising short-context tasks performance, eliminating the need for input-length-specific tuning, and preserving local positional information; and (b) avoiding attention logit outliers through attention logit interpolation rather than positional embedding interpolation.
We conducted extensive experiments across diverse long-context benchmarks and tasks, including LongBench\cite{bai-etal-2024-longbench}, L-Eval\cite{an-etal-2024-l}, and PG19\cite{Rae2019CompressiveTF}, demonstrating that GALI consistently outperforms existing state-of-the-art training-free methods. Furthermore, our analysis reveals a key insight: constraining interpolation to narrower positional intervals leads to improved performance, even on short-context tasks.


Main contributions are summarized as follows:
\begin{itemize}
\item We propose Greedy Attention Logit Interpolation (GALI),\textbf{ a training-free method that achieves superior and stable performance across both short- and long-context tasks without any input-length-specific tuning}. GALI integrates two key components: a greedy and localized position ID interpolation strategy, and a logit-level interpolation mechanism with Gaussian noise to simulate RoPE’s oscillatory behavior. These designs eliminate redundant extrapolation and attention logit outliers, enabling robust length extrapolation.

\item Our extensive evaluation on LongBench, L-Eval, and PG19 shows that GALI consistently outperforms existing training-free extrapolation methods. Further analysis reveals a key insight: \textbf{constraining extrapolation to narrower positional intervals improves performance even on short-context tasks}, emphasizing the importance of precise positional alignment in effective length extrapolation.
\end{itemize}

\section{RelatedWork}
\textbf{Rotary Position Embedding (RoPE): }
RoPE \cite{su2024roformer} is a technique that encodes positional information by applying rotary transformations to token embeddings, enabling relative position modeling in transformers. Given two token embeddings $\bm{x}_m, \bm{x}_n \in \mathbb{R}^l$ as query and key corresponding to position $m$ and $n$, the projection matrix $\bm{W}_Q, \bm{W}_K \in \mathbb{R}^{d \times l}$, RoPE applies a rotation to the projected token embeddings, i.e., $\bm{q}_m^{r}=(\bm{W}_Q\bm{x}_m)e^{im\bm{\theta}}$, $\bm{k}_n^{r}=(\bm{W}_K\bm{x}_n)e^{in\bm{\theta}}$, where $\bm{\theta}=[b^{0}, b^{-2/d}, \dots, b^{-2(j-1)/d}]$, $j \in [1,2, \dots, d/2]$ and $b$ is originally set to 10000. After that, the inner product between the query $\bm{q}_m^{r}$ and key $\bm{k}_n^{r}$ can be represented by the real part of ${\bm{q}_m^{r}}^*\bm{k}_n^{r}$, i.e.:

\vspace{-0.3cm}
{\small
\begin{flalign}
&\langle\bm{q}_m^{r},\bm{k}_n^{r}\rangle_{\mathbb{R}} = \operatorname{Re}(\langle(\bm{W}_Q\bm{x}_m)e^{im\bm{\theta}},(\bm{W}_K\bm{x}_m)e^{in\bm{\theta}}\rangle_{\mathbb{C}}) \nonumber\\
&= a(\bm{x}_m, \bm{x}_n, m-n)
\end{flalign}
}

$a(\cdot)$ is the function mapping token embeddings $\bm{x}_m, \bm{x}_n$ to the attention logit, which depends on their positional interval and is irrelevant to their absolute positions. Additionally, RoPE exhibits a long-term decay as positional interval increases \cite{su2024roformer}, as illustrated in Figure \ref{fig:app_logit_long_term_decay}.
Our proposed method, GALI, leverages two key properties of RoPE to achieve position interpolation and length extrapolation effectively.

\textbf{Positional Out-Of-Distribution (O.O.D.): }
In Transformer architectures, the self-attention mechanism is inherently position-agnostic, necessitating the use of position embeddings to encode positional information for processing ordered inputs \cite{Dufter2021PositionII, kazemnejad2024impact}. Even in large language models (LLMs) with causal attention, explicit positional encoding through position embeddings remains the standard approach.\footnote{Recent studies suggest causal attention implicitly encodes positional information, enabling performance without explicit position embeddings. However, this is beyond the scope of this paper.} During inference, when LLMs encounter input sequences exceeding the maximum length seen during training, the use of unseen position IDs causes a positional out-of-distribution (O.O.D.) issue, leading to degraded performance \cite{chen2023extending, Jin2024LLMML, xu2024base}. In the Rotary Position Embedding (RoPE) mechanism, extrapolating position IDs beyond the training range introduces untrained positional intervals, disrupting the attention score distribution. In contrast, position interpolation has yielded more stable attention distributions, requiring fewer fine-tuning steps. This observation has inspired subsequent interpolation-based methods \cite{chen2023clex, Xiong2023EffectiveLS, li2023functional, ding2024longrope, li2024extending, wu2024never}, as well as training-free approaches that map position interpolation into alternative frequency dimensions in embeddings \cite{LocalLLaMA-2023-ntk, LocalLLaMA-2023-dyn-ntk, peng2023yarn}. Recent work has explored other training-free length extrapolation techniques, such as group position IDs \cite{Jin2024LLMML} or chunk attention \cite{an2024training}.

\section{Method}

\begin{figure*}[t]
    \centering
    \includegraphics[width=0.8\linewidth]{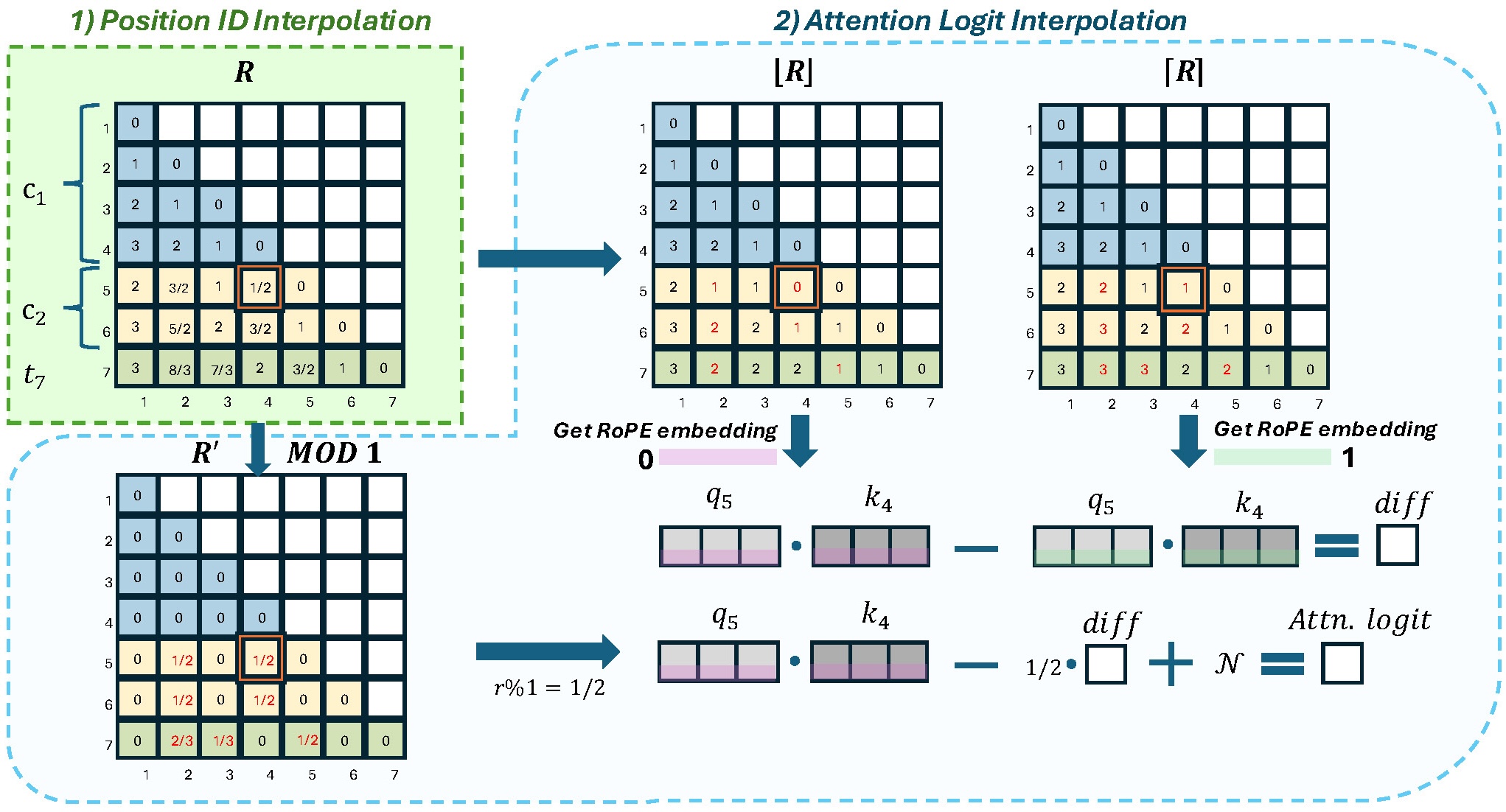}
    \caption{The overall procedure of the proposed GALI framework. The green dashed line illustrates \textbf{position ID interpolation}, while the blue dashed line shows \textbf{attention logit interpolation}. In this example, the training context window $L_{tr}$ is 4, the chunk size $s$ is 2, the local window $L_{w}$ is 2, and the prefill length is 6. Chunks are denoted as $c_1$ (first chunk) and $c_2$ (second chunk), while $t_7$ represents the first generated token. The positional interval matrix $R$ incorporates $\lceil R \rceil$, $\lfloor R \rfloor$, and $R^{\prime}$, representing ceiling, floor, and modulo operations, respectively. Red numbers in the positional interval matrix indicate interpolated positional intervals and $\mathcal{N}$ represents the Gaussian noise.}
    \label{fig:overall_method}
    \vspace{-3mm}
\end{figure*}

We introduces Greedy Attention Logit Interpolation (GALI), a novel training-free length extrapolation method that achieves superior and consistent performance across both short- and long-context tasks without requiring any input-length-specific tuning. GALI accomplishes this through two key mechanisms: (1) a greedy and localized interpolation strategy that preserves pretrained position IDs within the training context window and only interpolates beyond it when necessary, and (2) logit-level interpolation that avoids attention logit outliers by approximating attention logit between valid relative positions. To further stabilize attention behavior, GALI adds distance-scaled Gaussian noise that simulates the oscillatory nature of RoPE. The overall process is shown in Figure \ref{fig:overall_method}.


\subsection{Position ID Interpolation}
The proposed GALI introduces a greedy and localized interpolation strategy that minimizes deviation from pretrained positional distributions. Instead of applying global scaling across all positions, as done in NTK, Dyn-NTK, YaRN, or SelfExtend, GALI retains original position IDs within the training context window and interpolates only when necessary. This fine-grained, chunk / token-wise approach avoids redundant extrapolation and eliminates the sensitivity to input length observed in prior methods.

This strategy builds on insights from Dyn-NTK \cite{LocalLLaMA-2023-dyn-ntk}, which adjusts scaling factors based on the input length. However, Dyn-NTK still applies a global scaling factor to the entire sequence, overlooking that different tokens require different amounts of interpolation. For example, a token just beyond the training context window only needs one new position ID, while later tokens require more. Ideally, each token would have its own customized interpolation, but this is computationally expensive. GALI addresses this by grouping tokens into chunks and applying chunk-specific interpolation, avoiding global scaling.

Concretely, GALI segments the portion of the input beyond the training context window into fixed-size chunks when computing the positional interval matrix. Within each chunk, a local window of length $L_w$ preserves the original pretrained positional IDs. Only the remaining tokens are assigned interpolated IDs, determined by how many positions exceed the training context window, and computed to minimize disruption.

In the prefill stage, given an input sequence $S = (w_1, w_2, \dots, w_{L_{tr}}, \dots, w_{L})$ where $L_{tr}$ is the training context window size and $L$ is the input length in the prefill stage. We first divide it into chunks $C = (c_1, c_2, \dots, c_{L_{c}})$, where the size of $c_1$ is $L_{tr}$ 
 and other chunks have a size $s$, so the $L_c = \lceil {\frac{L-L_{tr}}{s}} \rceil + 1$. After that, we assign position IDs $S = (s_1, s_2, \dots, s_{L_{c}})$ for each chunk, where $s_1=[0,1,\dots,L_{tr}-1]$ and others are interpolated position IDs according to the following formula:

\vspace{-0.5cm}
{\small
\begin{align}
    \begin{cases}
        & j > 1; \\
        & {g}_j = \lceil {\frac{\sum_{i=1}^{j}{len(c_i)}-L_w}{L_{tr} - L_w}} \rceil;\\
        & {v}_j = 1/{g}_j; \\ 
        & s_j = [0, 1*{v}_j, 2*{v}_j, \dots, ({g}_j - 1) * {v}_j,\\
        & 1, \dots, L_{tr}-L_w-1, L_{tr}-L_w, \dots, L_{tr}-1]
    \end{cases}
    \label{equ:pid interpolation}
\end{align}
}

\textbf{Note that each chunk uses the complete pretrained positional intervals, making use of all the pretrained positional information greedily}.
During decoding, where tokens are generated sequentially, GALI applies the same greedy principle: each new token is treated as a single-token chunk, and its attended position IDs are interpolated based on how many attended tokens exceed the training context window.

Overall, this strategy ensures that each chunk or token reuses the full range of pretrained positional intervals when constructing the positional interval matrix. It avoids unnecessary positional distortion and removes the need for input-length-specific tuning required by global scaling methods. If the input length is within $L_{tr}$, no interpolation is applied; otherwise, the model preserves fidelity within the trained range and applies minimal-impact interpolation only to the extended portion.

\subsection{Attention Logit Interpolation}
To eliminate attention logit outliers and ensure robust extrapolation, GALI performs interpolation directly at the attention logit level, bypassing the need to compute position embeddings for unseen positional intervals. Unlike the methods that manipulate positional embeddings, which often produce unstable or extreme logit when extrapolated, GALI approximates logit via local linear interpolation and stabilizes them with Gaussian noise.
This design draws on observations about the behavior of RoPE: while it encodes relative positions with oscillatory trigonometric functions, these functions become numerically unstable when extrapolated beyond pretrained ranges. Instead of applying RoPE embeddings to interpolated positions, GALI interpolates between known attention logit corresponding to valid pretrained positional intervals.

Concretely, for two tokens $\bm{x}_m, \bm{x}_n \in \mathbb{R}^l$ at positions $m$ and $n$ (which may be floats due to interpolation), we define their positional interval as $r = m - n$. When $r$ is an integer, the corresponding attention logit is already trained and can be used directly. When $r$ is fractional, GALI linearly interpolates between the logit at $\lfloor r \rfloor$ and $\lceil r \rceil$, and introduces noise proportional to the positional interval to preserve oscillatory behavior:

\vspace{-0.3cm}
{\small
\begin{flalign}
&a(\bm{x}_m, \bm{x}_n, r) = a(\bm{x}_m, \bm{x}_n,\lfloor{r}\rfloor) - [a(\bm{x}_m, \bm{x}_n, \lfloor{r}\rfloor)\\
&-a(\bm{x}_m, \bm{x}_n, \lceil{r}\rceil)] \nonumber *({r}{\%}1)+\mathcal{N}(0, {\frac{r}{L_{tr}}}^2)
\label{equ:attn logit interpolation}
\end{flalign}
}
\vspace{-0.3cm}

To enable efficient matrix operations with the positional interval matrix $R$ in the computation process (Figure~\ref{fig:overall_method}), we employ an approximate implementation by substituting $r$ with $r=\lceil{m}\rceil - n$, as elaborated in the pseudo-code in Appendix~\ref{app: gali pseudo code}.

By avoiding embedding-level extrapolation and operating directly on logit, GALI eliminates outliers and achieves robust length extrapolation over long sequences in a training-free manner.

\section{Experiments}
We evaluate GALI on Llama3-8B-ins models across two task categories: real-world long-context tasks and long-context language modeling tasks. For comparison, we implement all published training-free length extrapolation methods, including NTK\cite{LocalLLaMA-2023-ntk}, Dyn-NTK\cite{LocalLLaMA-2023-dyn-ntk}, YARN\cite{peng2023yarn}, SelfExtend\cite{Jin2024LLMML}, and ChunkLlama\cite{an2024training}. Appendix \ref{app: data stastics} details data statistics.

\subsection{Experiments Setup}
\textbf{Real-world long-context task: } We evaluate GALI on two widely used long-context benchmarks, LongBench\cite{bai-etal-2024-longbench} and L-Eval\cite{an-etal-2024-l}. For LongBench, we use 16 English datasets, while for L-Eval, we focus on closed-ended groups. For consistency, we follow the official task prompt templates and truncation strategies from the respective benchmarks. 



\textbf{Long-context language modeling task: } To evaluate GALI's long-context language modeling capabilities, we use the test split of PG19\cite{Rae2019CompressiveTF}, an open-vocabulary language modeling benchmark derived from Project Gutenberg.

\begin{table*}[t]
\fontsize{18}{24}\selectfont
\setlength{\tabcolsep}{5pt}
\centering
\resizebox{\textwidth}{!}{
\begin{tabular}{c|cccccccccccccccccc}
\specialrule{1pt}{0pt}{2pt}
\toprule
 &\multirow{4}{*}{~~~~~~~~~~~~~\textbf{Methods}~~~~~~~~~~~~~} & \multicolumn{3}{c}{\textbf{Single document QA}} & \multicolumn{3}{c}{\textbf{Multi document QA}} & \multicolumn{3}{c}{\textbf{Summarization}} & \multicolumn{3}{c}{\textbf{Few-shot Learning}} & \multicolumn{2}{c}{\textbf{Synthetic}} & \multicolumn{2}{c}{\textbf{Code}} & \multirow{4}{*}{\textbf{Average}} \\
\cmidrule(r){3-5} \cmidrule(r){6-8} \cmidrule(r){9-11} \cmidrule(r){12-14} \cmidrule(r){15-16} \cmidrule(r){17-18}
& & \rotatebox{30}{NarrativeQA} & \rotatebox{30}{Qasper} & \rotatebox{30}{MultiField-en} & \rotatebox{30}{HotpotQA} & \rotatebox{30}{2WikiMQA} & \rotatebox{30}{Musique} & \rotatebox{30}{GovReport} & \rotatebox{30}{QMSum} & \rotatebox{30}{MultiNews} & \rotatebox{30}{TREC} & \rotatebox{30}{TriviaQA} & \rotatebox{30}{SAMSum} & \rotatebox{30}{PassageCount} & \rotatebox{30}{PassageRe} & \rotatebox{30}{Lcc} & \rotatebox{30}{RepoBench-P} & \\

\midrule

\multirow{7}{*}{\rotatebox[origin=c]{90}{\fontsize{18}{100}\selectfont Llama3-8b-ins-4k}}  &  Original  &  17.83  &  40.62  &  47.02  &  40.97  &  35.15  &  20.99  &  27.76  &  19.70  &  24.62  &  71.00  &  89.54  &  42.31  &  6.00  &  23.50  &  56.96  &  49.06  &  \cellcolor{YellowGreen!25}{38.31} \\
 &  SelfExtend-16k  & 23.34 & 44.59 & 51.22 & 44.91 & 37.43 & 29.50 & 28.52 & 22.14 & 24.34 & 75.50 & 90.71 & 42.58 & 7.50 & 92.50 & 54.99 & 50.83 & \cellcolor{YellowGreen!25}{45.04} \\
 &  ChunkLlama-16k & 20.91 & 40.15 & 49.87 &  47.71  &  \textbf{40.80}  & 28.75 & 30.37 & 21.81 & 24.32 & 74.50 & 90.29 & 41.78 & 2.50 & 56.75 &  \textbf{58.99}  &  \textbf{57.55}  & \cellcolor{YellowGreen!25}{42.94} \\
 &  NTK-16k  &  22.59  &  \textbf{46.25}  & 53.21 &  \textbf{51.91}  & 37.51 & 26.56  &  \textbf{30.69}  &  \textbf{22.74}  & 24.03 & 73.50 & 90.46 & 42.20  &  \textbf{11.50}  & 73.00 & 34.53 & 36.39 & \cellcolor{YellowGreen!25}{42.32} \\
 &  Dyn-NTK-16k & 18.65 & 44.91 & 51.37 & 46.28 & 37.57 & 28.03 & 30.20 & 21.53 & 24.48  &  76.00  &  89.11  &  42.88  &  9.00  & 74.50  & 53.91 & 32.65 & \cellcolor{YellowGreen!25}{42.57} \\
 &  YaRN-16k & 16.43 & 40.13  &  \textbf{53.04}  & 45.93 & 33.66 & 28.51 & 30.40 & 22.42 & 23.24 & 75.50 &  91.04  &  \textbf{44.53}  & 6.50 & 86.50 & 43.26 & 48.26 & \cellcolor{YellowGreen!25}{43.08} \\
  & \cellcolor{WildStrawberry!25}\textbf{(Ours)GALI-16k}  &  \cellcolor{WildStrawberry!25}\textbf{24.69}  &  \cellcolor{WildStrawberry!25}45.26  &  \cellcolor{WildStrawberry!25}51.78  &  \cellcolor{WildStrawberry!25}51.33  &  \cellcolor{WildStrawberry!25}37.16  &  \cellcolor{WildStrawberry!25}\textbf{30.79}  &  \cellcolor{WildStrawberry!25}29.28  &  \cellcolor{WildStrawberry!25}22.65  &  \cellcolor{WildStrawberry!25}\textbf{24.63}  &  \cellcolor{WildStrawberry!25}\textbf{77.00}  &  \cellcolor{WildStrawberry!25}\textbf{91.61}  &  \cellcolor{WildStrawberry!25}42.92  &  \cellcolor{WildStrawberry!25}9.00  &  \cellcolor{WildStrawberry!25}\textbf{95.5}  &  \cellcolor{WildStrawberry!25}56.84  &  \cellcolor{WildStrawberry!25}49.04  &  \cellcolor{YellowGreen!25}{\textbf{46.22}} \\
\midrule

\multirow{8}{*}{\rotatebox[origin=c]{90}{\fontsize{18}{100}\selectfont Llama3-8b-ins-8k}}  &  
Original\color{blue}\textsuperscript{*} & 21.71 & 44.24 & 44.54 & 46.82 & 36.42 & 21.49 & 30.03 & 22.67 & \textbf{27.79} & 74.50 & 90.23 & \textbf{42.53} &  0.00  &  67.00  &  57.00  & 51.22 & \cellcolor{YellowGreen!25}{42.39} \\
 &  SelfExtend-16k\color{blue}\textsuperscript{*} & 21.50 & 43.96 & \textbf{50.26} & 48.18 & 28.18 & 25.58 & \textbf{34.88} & \textbf{23.83} & 26.96 & 75.50 & 88.26 & 42.01 & 4.12 & 88.00 & 36.58 & 37.73 & \cellcolor{YellowGreen!25}{42.22} \\
 &  ChunkLlama-16k & 23.87 & 43.86 & 46.97 & 49.37 & 35.34 & 26.52 & 31.06 & 21.99 & 24.45 & 76.00 & 90.73 & 42.29 & \textbf{7.00} & 72.00 & \textbf{59.93} & \textbf{56.98} & \cellcolor{YellowGreen!25}{44.27} \\
 &  NTK-16k & 8.04 & 43.85 & 47.94 & 20.44 & 34.32 & 1.57 & 24.31 & 13.22 & 24.12 & 74.50 & 52.18 & 33.12 & 4.50 & 45.50 & 46.84 & 38.71 & \cellcolor{YellowGreen!25}{32.07} \\
 &  Dyn-NTK-16k & 8.19 & 43.31 & 47.91 & 34.63 & 35.26 & 7.92 & 26.83 & 17.85 & 24.51 & 76.50 & 71.72 & 39.15 & 5.67 & 83.50 & 56.58 & 46.39 & \cellcolor{YellowGreen!25}{39.12} \\
 &  YaRN-16k & 12.39 & 42.60 & 51.70 & 40.06 & 35.03 & 12.81 & 30.30 & 22.56 & 23.51 & 75.50 & 82.99 & 42.31 & 6.50 & \textbf{89.00} & 50.51 & 51.58 & \cellcolor{YellowGreen!25}{41.83} \\
 &  \cellcolor{WildStrawberry!25} \textbf{(Ours)GALI-16k}  &  \cellcolor{WildStrawberry!25}\textbf{25.88}  &  \cellcolor{WildStrawberry!25}\textbf{45.65}  &  \cellcolor{WildStrawberry!25}47.09  &  \cellcolor{WildStrawberry!25}\textbf{51.07}  &  \cellcolor{WildStrawberry!25}\textbf{37.42}  &  \cellcolor{WildStrawberry!25}\textbf{28.75}  &  \cellcolor{WildStrawberry!25}30.09  &  \cellcolor{WildStrawberry!25}22.7  &  \cellcolor{WildStrawberry!25}24.58  &  \cellcolor{WildStrawberry!25}\textbf{77.00}  &  \cellcolor{WildStrawberry!25}\textbf{90.91}  &  \cellcolor{WildStrawberry!25}42.43  &  \cellcolor{WildStrawberry!25}6.00  &  \cellcolor{WildStrawberry!25}83.00  &  \cellcolor{WildStrawberry!25}57.04  &  \cellcolor{WildStrawberry!25}53.06  & \cellcolor{YellowGreen!25}{\textbf{45.17}} \\

\midrule

\multirow{8}{*}{\rotatebox[origin=c]{90}{\fontsize{18}{100}\selectfont Llama3-8b-ins-8k}}  &  Original\color{blue}\textsuperscript{*} & 21.71 & 44.24 & 44.54 & 46.82 & 36.42 & 21.49 & 30.03 & 22.67 & \textbf{27.79} & 74.50 & 90.23 & 42.53 & 0.00 & 67.00 & 57.00 & 51.22 & \cellcolor{YellowGreen!25}{42.39} \\
 &  SelfExtend-32k\color{blue}\textsuperscript{*} & 12.04 & 12.10 & 20.15 & 8.22 & 9.68 & 3.89 & 27.90 & 14.58 & 22.13 & 61.00 & 82.82 & 1.40 & 2.37 & 2.83 & 57.87 & 56.42 & \cellcolor{YellowGreen!25}{24.71} \\
 &  SelfExtend-32k & 26.27 & 44.23 & 50.19 & 48.28 & 38.29 & 29.19 & 29.24 & 22.68 & 24.59 & 76.00 & 90.16 & 42.45 & 8.00 & 88.00 & 57.47 & 49.51 & \cellcolor{YellowGreen!25}{45.28} \\
 &  ChunkLlama-32k & 24.48 & 42.37 & 47.05 & 48.79 & 34.53 & 26.94 & \textbf{32.08} & \textbf{23.40} & 24.36 & 76.00 & 90.46 & 42.08 & 6.50 & 72.00 & \textbf{59.52} & \textbf{60.54} & \cellcolor{YellowGreen!25}{44.44} \\
 &  NTK-32k & 7.31 & 45.11 & \textbf{53.18} & 52.31 & 37.70 & 27.37 & 29.37 & 21.45 & 23.69 & 73.50 & 78.25 & 41.83 & \textbf{9.00} & 69.00 & 34.25 & 36.12 & \cellcolor{YellowGreen!25}{39.97} \\
 &  Dyn-NTK-32k & 23.06 & 43.95 & 48.55 & \textbf{52.68} & 37.46 & 25.22 & 31.53 & 22.19 & 24.52 & \textbf{77.00} & 90.96 & 42.42 & 8.00 & 71.50 & 56.77 & 43.78 & \cellcolor{YellowGreen!25}{43.72} \\
 &  YaRN-32k & 17.09 & 40.90 & 52.51 & 46.40 & 33.92 & \textbf{29.47} & 29.93 & 22.69 & 23.11 & 75.00 & \textbf{91.29} & \textbf{42.54} & 5.50 & \textbf{89.50} & 46.50 & 51.38 & \cellcolor{YellowGreen!25}{43.61} \\
 &  \cellcolor{WildStrawberry!25}\textbf{(Ours)GALI-32k}  & \cellcolor{WildStrawberry!25}\textbf{28.63}  &  \cellcolor{WildStrawberry!25}\textbf{45.66}  &  \cellcolor{WildStrawberry!25}47.23  &  \cellcolor{WildStrawberry!25}51.07  &  \cellcolor{WildStrawberry!25}\textbf{38.35}  &  \cellcolor{WildStrawberry!25}29.00  &  \cellcolor{WildStrawberry!25}29.98  &  \cellcolor{WildStrawberry!25}22.79  &  \cellcolor{WildStrawberry!25}24.59  &  \cellcolor{WildStrawberry!25}\textbf{77.00}  &  \cellcolor{WildStrawberry!25}91.13  &  \cellcolor{WildStrawberry!25}42.38  &  \cellcolor{WildStrawberry!25}5.50  &  \cellcolor{WildStrawberry!25}83.00  &  \cellcolor{WildStrawberry!25}57.07  &  \cellcolor{WildStrawberry!25}52.63 & \cellcolor{YellowGreen!25}{\textbf{45.38}} \\



\bottomrule
\end{tabular}
}
\caption{Performance comparison on LongBench. The best result in each experiment is bolded. Results marked with {\color{blue}\textsuperscript{*}} are reported by LongBench \cite{Jin2024LLMML}. The number following each method denotes the target context window size (e.g., 16k represents 16 × 1024 tokens). "Original" refers to evaluations conducted using the backbone model in the left column. Additional results using the Llama2-7B-Chat-4K backbone are provided in Appendix \ref{app:real world task}.
}
\label{tab:longbench_all}
\end{table*}

\textbf{Backbone models and baseline methods: } We use Llama3-8b-ins-4k (Llama3-4k) and Llama3-8b-ins-8k (Llama3-8k) as backbone models, where the number following each model indicates its initial context window size. We obtain Llama3-4k backbone via modifying its max\_position\_embedding parameter. We use shorter-versions of Llama3-8b-ins over other LLMs with shorter training context windows like LLama2 since it cannot fully understand all pretrained positional intervals, which limits GALI’s effectiveness in practice. The effective understanding range of LLMs is shorter than their training context window, as evidenced in \cite{Jin2024LLMML, Hsieh2024RULERWT}. For the baseline methods, we compare with NTK\cite{LocalLLaMA-2023-ntk}, Dyn-NTK\cite{LocalLLaMA-2023-dyn-ntk}, YaRN\cite{peng2023yarn} using huggingface implementation and SelfExtend\cite{Jin2024LLMML}, ChunkLlama\cite{an2024training} with their official implementation. They are all of the training-free length extrapolation methods up to now. The implementation details of these methods can be found in Appendix \ref{app: imp. details}.

\subsection{Real-World long-context Task Results}
\label{main:Real-world long-context task results}
The LongBench results (Table \ref{tab:longbench_all}) highlight GALI’s strong average performance on the Llama3-8b-ins backbone series, surpassing both the 4k and 8k backbones as well as all other methods. Notably, (1) when using the Llama3-8k backbone with a 32k context window, GALI's average score improves only slightly over the 16k setting (by 0.21), whereas other methods—except ChunkLlama—achieve much larger gains (often exceeding 1 point ); (2) using a 16k context window on the Llama3-4k backbone yields better results than the same context window on Llama3-8k. Again, GALI exhibits only minor gains in this setting, while other methods show substantially larger improvements. These two observations demonstrate that: (1) GALI achieves stable and superior performance without requiring input-length-specific tuning; and (2) under current LLM architectures, performing extrapolation within a narrower positional interval range leads to better results, even on short-context tasks.

First, we note that Figure \ref{fig:llama3_length_dist} shows most LongBench samples are shorter than 16k tokens when tokenized by Llama3. However, NTK, Dyn-NTK, YaRN, and SelfExtend all benefit significantly from expanding the context window to 32k using the Llama3-8k backbone, especially on HotpotQA and Musique\footnote{SelfExtend is highly sensitive to its hyperparameters, as noted in its GitHub. So, our reproduced results on Llama3-8k at 32k far exceed those reported in the original paper.}, whose most samples fall below 16k tokens. This highlights a key weakness of global scaling methods: they require tuning of the global scaling factor according to input length. Moreover, simply matching the target context window to the input length is not sufficient, because misalignment in positional mapping can still lead to significant performance drops. In contrast, GALI improves by 2.75 on the longest dataset NarrativeQA,  when moving from 16k to 32k, while performance on sequences shorter than 16k remains nearly unchanged. This confirms that GALI can achieve stable and superior results without any input-length-specific tuning. Meanwhile, ChunkLlama exhibits only minor fluctuations. These results reflect the fundamental differences between approaches:

GALI maps inputs to the full range of positional intervals learned in pretraining, while NTK, Dyn-NTK, YaRN, and SelfExtend map inputs into a fixed range based on the input length and the global scaling factor (e.g., in SelfExtend, a larger group size leads to a narrower range). ChunkLlama determines its effective positional mapping through hyperparameters like chunk size and local window. 

\begin{figure}[htp]
    \centering
    \subfigure[LongBench]{
        \includegraphics[width=0.45\linewidth, trim=0 0 15 0, clip]{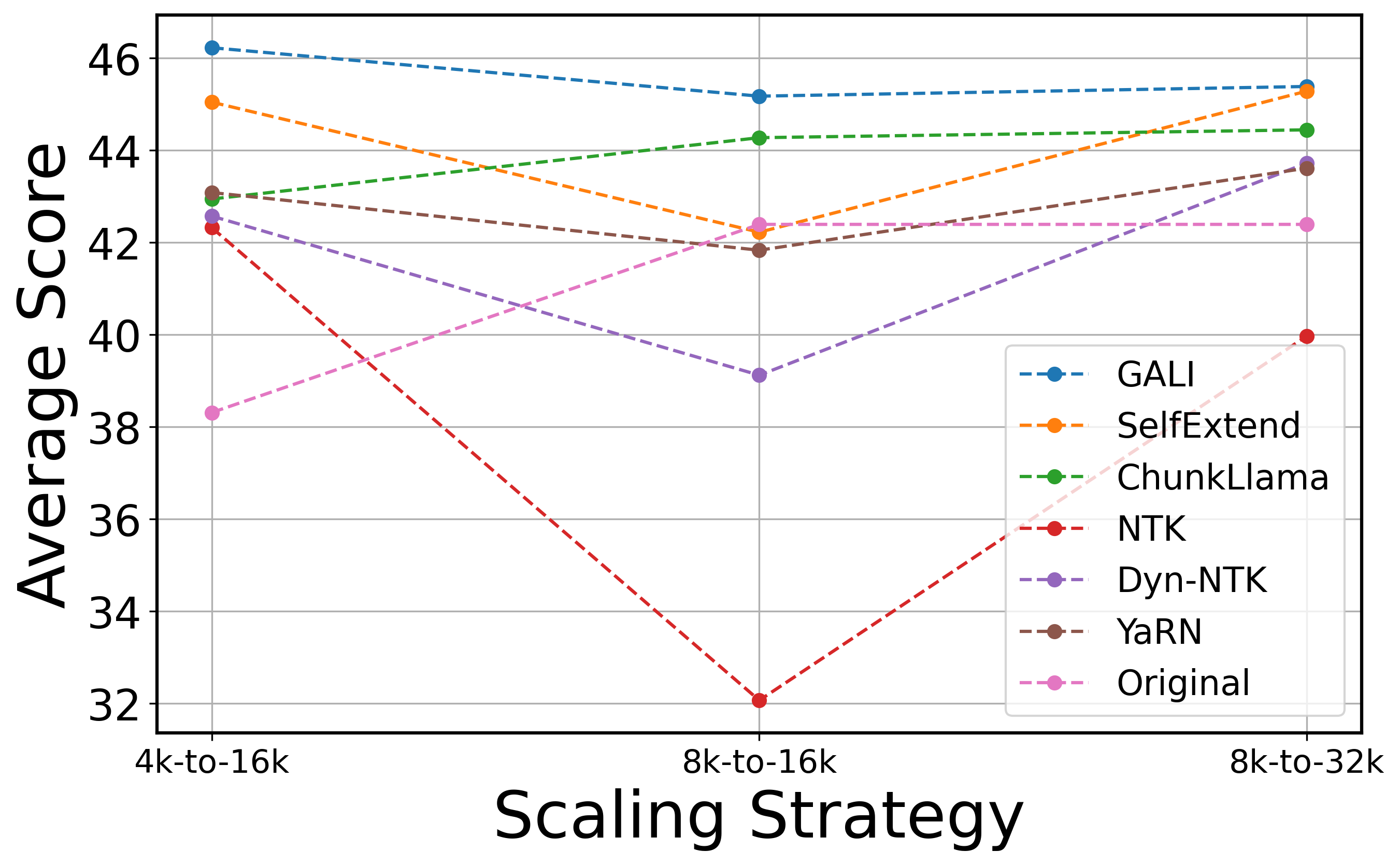}
        \label{fig:lbench_trend}
    }
    \subfigure[L-Eval]{
        \includegraphics[width=0.45\linewidth,trim=0 0 15 0, clip]{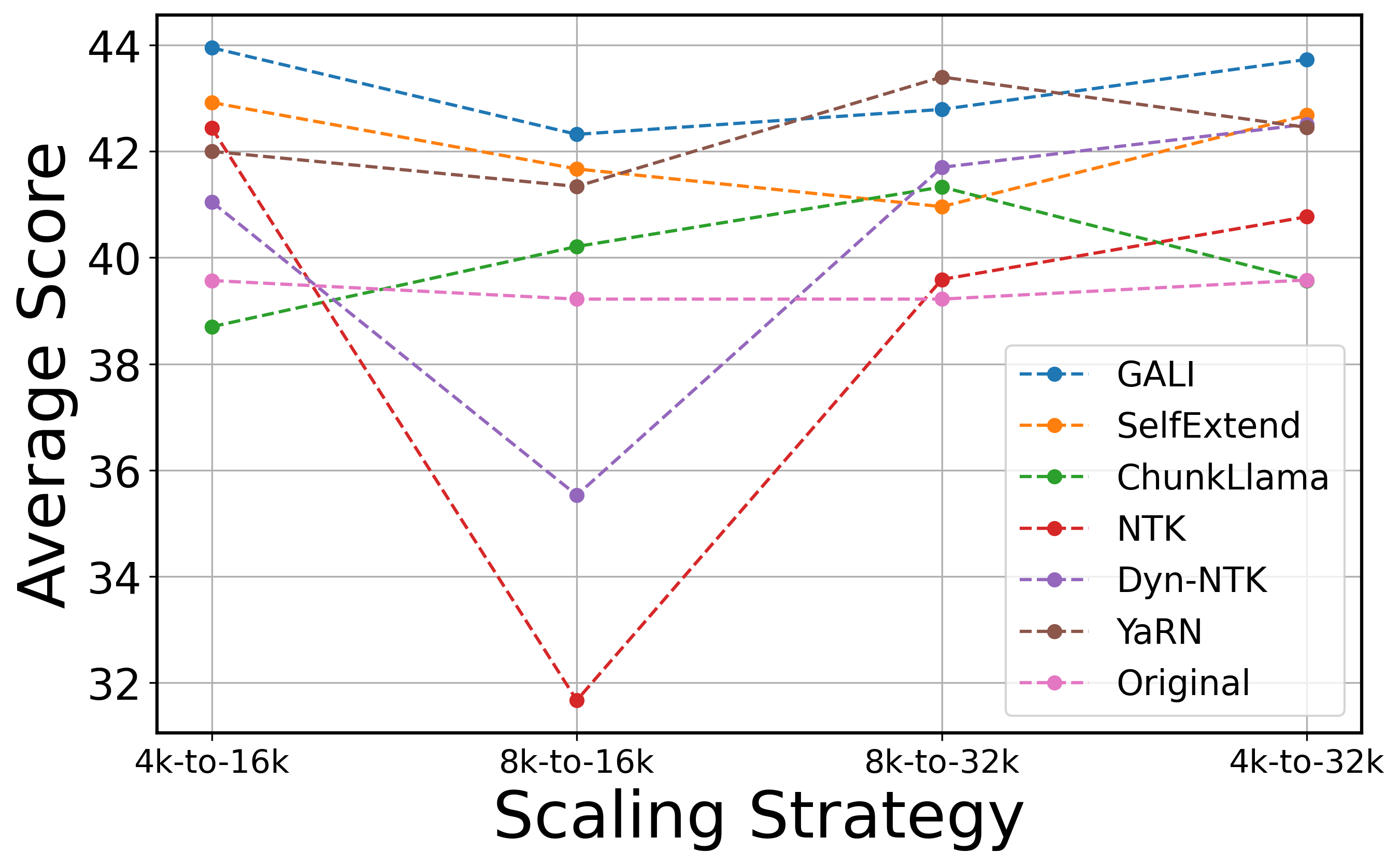}
        \label{fig:leval_trend}
    }
    \caption{The trend of average scores across different methods and settings. The X-axis tick “4k-to-16k” represents an initial context window of 4k and a target window of 16k. For a more meaningful comparison, the average scores are computed on the three long-text datasets in L-Eval: Coursera, SFiction, and CodeU.}
    \label{fig:real_world_task_trend}
\end{figure}

Consider HotpotQA and Musique: a 32k context window on the Llama3-8k maps 16k-length inputs to [0, 4096), while a 16k window maps to [0, 8192). For SelfExtend, this mapping is not fixed and depends on its hyperparameters, but typically compresses the range. Thus, even though a 16k window is sufficient to cover the input, NTK, Dyn-NTK, YaRN, and SelfExtend still benefit from using 32k, as it places more tokens in narrower positional ranges. This implies that these methods are sensitive to the scaling factor, making it difficult to determine optimal settings in practice. GALI, by contrast, achieves stable and superior performance without such tuning. Although ChunkLlama also avoids scaling, it suffers from degraded performance due to the loss of local positional information.
Second, it is important to recognize that LLMs interpret positional intervals differently depending on their training context window, as observed in \cite{Hsieh2024RULERWT}. This explains why a 16k window on Llama3-4k outperforms the same 16k setting on Llama3-8k: since LLMs are trained via next-token prediction, they are more familiar with shorter positional intervals. 
As a result, all methods except ChunkLlama perform better with a 16k context window on Llama3-4k than with the same context window on Llama3-8k.
The L-Eval results further support our analysis. As shown in Table \ref{tab:leval_all}, GALI achieves the highest average performance across most configurations, except when using a 32k context window on Llama3-8k. This is consistent with the LongBench pattern: while 32k on Llama3-8k performs slightly better than 16k, it still falls short of 16k on Llama3-4k, again reflecting how narrower positional ranges improve extrapolation performance. Since Llama3 better understands shorter intervals, methods that remap text into a smaller positional range gain an advantage at 32k, while GALI's full-span reuse may be less aligned in this case. However, when we tested GALI with a 32k context window on Llama3-4k to force it to operate within the [0, 4096) interval, it once again achieved the best results. Figure \ref{fig:leval_trend} summarizes these performance trends.
Additionally, GSM, QuALITY, and TOEFL, whose inputs remain below 8k with the Llama3 tokenizer, show consistent gains for SE, YaRN, and NTK over the base model, as shown in Appendix \ref{app:real world task}. These results confirm that mapping into narrower, well-trained positional intervals benefits extrapolation, even for short context tasks.
Experiments across LongBench and L-Eval support two key conclusions: (1) GALI avoids global scaling, requires no input-length-specific tuning, and achieves superior and stable performance through logit-level interpolation and greedy reuse of pretrained intervals; and(2) Training-free extrapolation methods benefit from using narrower positional intervals, compensating for LLMs’ uneven positional understanding.

\begin{table}[t]
\fontsize{18}{24}\selectfont
\setlength{\tabcolsep}{5pt}
\centering
\caption{Performance comparison on L-Eval. The best results are bolded. The GSM, QuALITY, and TOEFL were excluded here since their sequence lengths remain below 8192 tokens when using the Llama3 tokenizer, making them unsuitable for long-context evaluation. Results for these three datasets, along with those for all datasets using the Llama2-7B-Chat-4K (Llama2-4k) backbone, are provided in Appendix \ref{app:real world task}.}
\resizebox{0.49\textwidth}{!}{
\begin{tabular}{c|cccccccc}
\specialrule{1pt}{0pt}{2pt}
\toprule
 & ~Methods~ & Coursera & GSM & QuALITY& TOFEL & SFiction & CodeU & Average \\



\midrule

\multirow{7}{*}{\rotatebox[origin=c]{90}{\fontsize{18}{100}\selectfont Llama3-8b-ins-4k}} &  Original &53.34 & 75.00 & 59.41 & 81.41 & 60.94 & 4.44 & \cellcolor{YellowGreen!25} 55.76  \\
 &  SelfExtend-16k & 55.23 & 79.00 & 64.36 & 79.18 & \textbf{67.97} & 5.56 & \cellcolor{YellowGreen!25}58.55 \\
 &  ChunkLlama-16k & 52.62 & 77.00 & 63.37 & 81.04 & 60.16 & 3.33 & \cellcolor{YellowGreen!25}56.25 \\
 &  NTK-16k & \textbf{57.70} & 80.00 & 63.86 & 81.04 & 64.06 & 5.56 & \cellcolor{YellowGreen!25}58.70 \\
 &  Dyn-NTK-16k & 54.07 & 75.00 & 64.36 & 82.16 & \textbf{67.97} & 1.11 & \cellcolor{YellowGreen!25}57.44 \\
 &  YaRN-16k & 56.40 & \textbf{81.00} & 59.40 & 79.18 & 64.06 & 5.56 & \cellcolor{YellowGreen!25}57.60 \\
 &  \cellcolor{WildStrawberry!25}\textbf{(Ours)GALI-16k} & \cellcolor{WildStrawberry!25}56.54 & \cellcolor{WildStrawberry!25}74.00 & \cellcolor{WildStrawberry!25}\textbf{65.35} & \cellcolor{WildStrawberry!25}\textbf{84.06} & \cellcolor{WildStrawberry!25}66.41 & \cellcolor{WildStrawberry!25}\textbf{8.89} & \cellcolor{YellowGreen!25}\textbf{59.21} \\

\midrule

\multirow{7}{*}{\rotatebox[origin=c]{90}{\fontsize{18}{100}\selectfont Llama3-8b-ins-8k}} & 
Original & 53.05 & - & - & - & 60.16 & 4.44 & \cellcolor{YellowGreen!25}39.22 \\
& SelfExtend-16k & 55.38 & - & - & - & 64.06 & 5.56 & \cellcolor{YellowGreen!25}41.67 \\
& ChunkLlama-16k & 53.34 & - & - & - & 61.72 & 5.56 & \cellcolor{YellowGreen!25}40.21 \\
& NTK-16k & 52.03 & - & - & - & 42.97 & 0.00 & \cellcolor{YellowGreen!25}31.67 \\
& Dyn-NTK-16k & 52.03 & - & - & - & 52.34 & 2.22 & \cellcolor{YellowGreen!25}35.53 \\
& YaRN-16k & \textbf{55.96} & - & - & - & 62.5 & 5.56 & \cellcolor{YellowGreen!25}41.34 \\
& \cellcolor{WildStrawberry!25}\textbf{(Ours)GALI-16k} & \cellcolor{WildStrawberry!25}54.65 & \cellcolor{WildStrawberry!25}- & \cellcolor{WildStrawberry!25}- & \cellcolor{WildStrawberry!25}- & \cellcolor{WildStrawberry!25}\textbf{65.63} & \cellcolor{WildStrawberry!25}\textbf{6.67} & \cellcolor{YellowGreen!25}\textbf{42.32} \\

\midrule

\multirow{7}{*}{\rotatebox[origin=c]{90}{\fontsize{18}{100}\selectfont Llama3-8b-ins-4k}} &  Original &53.34 & 75.00 & 59.41 & 81.41 & 60.94 & 4.44 & \cellcolor{YellowGreen!25} 55.76  \\
& SelfExtend-32k & \textbf{54.51} & 80.00 & 64.36 & 77.70 & 67.97 & 5.56 & \cellcolor{YellowGreen!25} 58.35 \\
& ChunkLlama-32k & 53.20 & 75.00 & 63.37 & 81.04 & 63.28 & 2.22 & \cellcolor{YellowGreen!25} 56.35 \\
& NTK-32k & 52.91 & \textbf{82.00} & 61.39 & 79.93 & 67.19 & 2.22 & \cellcolor{YellowGreen!25} 57.60 \\
& Dyn-NTK-32k & 52.33 & 76.00 & 63.86 & 82.16 & \textbf{71.88} & 3.33 & \cellcolor{YellowGreen!25} 58.26 \\
& YaRN-32k & 53.05 & 73.00 & 59.41 & 79.55 & 68.75 & 5.56 & \cellcolor{YellowGreen!25} 56.55 \\
& \cellcolor{WildStrawberry!25}\textbf{(Ours)GALI-32k} & \cellcolor{WildStrawberry!25}54.17 & \cellcolor{WildStrawberry!25}74.00 & \cellcolor{WildStrawberry!25}\textbf{65.35} & \cellcolor{WildStrawberry!25}\textbf{84.06} & \cellcolor{WildStrawberry!25}68.75 & \cellcolor{WildStrawberry!25}\textbf{7.78} & \cellcolor{YellowGreen!25} \textbf{59.10} \\

\midrule

\multirow{7}{*}{\rotatebox[origin=c]{90}{\fontsize{18}{100}\selectfont Llama3-8b-ins-8k}} & 
Original & 53.05 & - & - & - & 60.16 & 4.44 & \cellcolor{YellowGreen!25}39.22 \\
& SelfExtend-32k & 53.92 & - & - & - & 65.63 & 3.33 & \cellcolor{YellowGreen!25}40.96 \\
& ChunkLlama-32k & 54.36 & - & - & - & 64.06 & 5.56 & \cellcolor{YellowGreen!25}41.33 \\
& NTK-32k & \textbf{58.28} & - & - & - & 59.38 & 1.11 & \cellcolor{YellowGreen!25}39.59 \\
& Dyn-NTK-32k & 54.36 & - & - & - & 64.06 & 6.67 & \cellcolor{YellowGreen!25}41.70 \\
& YaRN-32k & 55.23 & - & - & - & \textbf{67.19} & \textbf{7.78} & \cellcolor{YellowGreen!25}\textbf{43.40} \\
& \cellcolor{WildStrawberry!25}\textbf{(Ours)GALI-32k} & \cellcolor{WildStrawberry!25}54.17 & \cellcolor{WildStrawberry!25}- & \cellcolor{WildStrawberry!25}- & \cellcolor{WildStrawberry!25}- & \cellcolor{WildStrawberry!25}66.41 & \cellcolor{WildStrawberry!25}\textbf{7.78} & \cellcolor{YellowGreen!25}42.79 \\

\bottomrule
\end{tabular}
}


\label{tab:leval_all}
\end{table}

\begin{table}[thp]
\fontsize{18}{24}\selectfont
\setlength{\tabcolsep}{5pt}
\centering
\caption{Performance on the PG19 dataset across varying target context window sizes.}
\label{tab:pg19_models}
\resizebox{0.49\textwidth}{!}{
\begin{tabular}{c|cccccccccccc}
\specialrule{1pt}{0pt}{2pt}
\toprule
& \textbf{Methods} & \textbf{1k} & \textbf{4k} & \textbf{8k} & \textbf{12k} & \textbf{16k} & \textbf{20k} & \textbf{24k} & \textbf{28k} & \textbf{32k} \\ \midrule
\multirow{6}{*}{\rotatebox[origin=c]{90}{\fontsize{18}{100}\selectfont Llama3-8b-ins-8k}} & SelfExtend & 11.52 & 11.54 & 11.32 & 11.18 & 11.07 & 10.97 & 11.01 & 11.04 & 10.91 \\ 
& ChunkLlama & 11.72 & 11.77 & 11.54 & 11.39 & 11.27 & - & - & - & - \\ 
& NTK  & 11.93 & 11.94 & 11.67 & 11.50 & 11.39 & 13.03 & 23.00 & 42.95 & 77.41 \\ 
& Dyn-NTK  & 11.51 & 11.53 & 12.75 & 66.88 & 166.86 & 269.93 & 334.83 & 360.57 & 365.36 \\ 
& YaRN  & 11.93 & 11.81 & 11.48 & 11.30 & 11.18 & 11.06 & 11.10 & 11.13 & 11.18 \\ 
& \cellcolor{WildStrawberry!25}\textbf{(Ours)GALI} & \cellcolor{WildStrawberry!25}11.52 & \cellcolor{WildStrawberry!25}11.54 & \cellcolor{WildStrawberry!25}11.35 & \cellcolor{WildStrawberry!25}11.25 & \cellcolor{WildStrawberry!25}11.17 & \cellcolor{WildStrawberry!25}11.09 & \cellcolor{WildStrawberry!25}11.14 & \cellcolor{WildStrawberry!25}11.18 & \cellcolor{WildStrawberry!25}11.05 \\ 
\bottomrule
\end{tabular}
}
\end{table}

\subsection{Long Language Modeling Task Results}
The language modeling results are shown in Table \ref{tab:pg19_models}. Due to OOM, we cannot get ChunkLlama's PPL results when setting the maximum position embedding to 32768. Except for NTK and DYN-NTK, all methods maintained a stable PPL without exploding. While low PPL does not guarantee better real-world task performance, an exploding PPL is a clear indicator of performance degradation in downstream tasks. Notably, GALI achieved the second-lowest PPL, demonstrating superior stability in length extrapolation. We tested PPL using a 16k contest window with Llama2-4k backbone. Please refer to the Appendix \ref{app:ppl_test}.

\begin{figure*}[htp]
    \centering
    \subfigure[Attention Score Matrix Differences]{
        \includegraphics[width=0.31\linewidth]{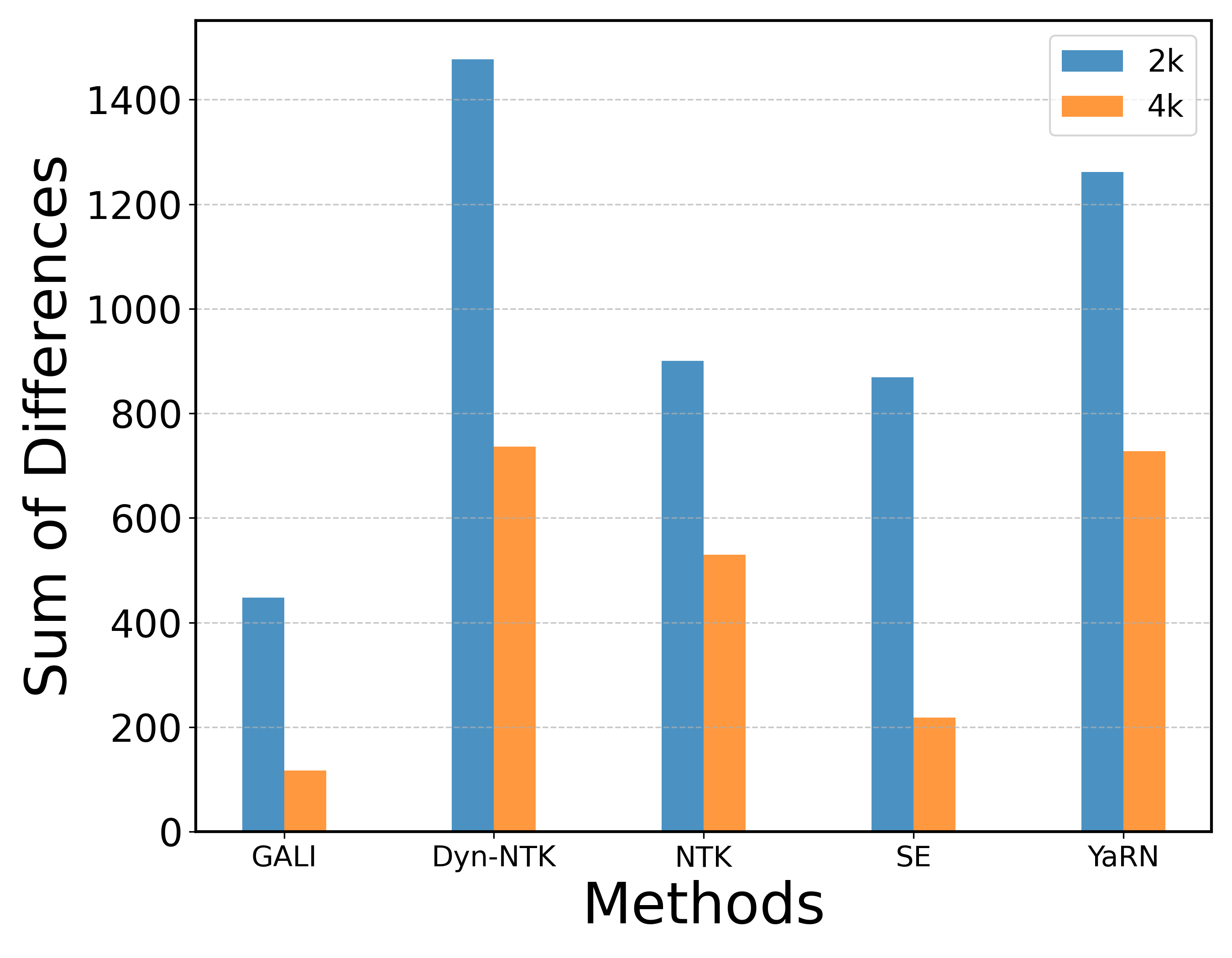}
        \label{fig:attn_differences_grouped}
    }
    \subfigure[Row Entropy Differences (2K)]{
        \includegraphics[width=0.31\linewidth]{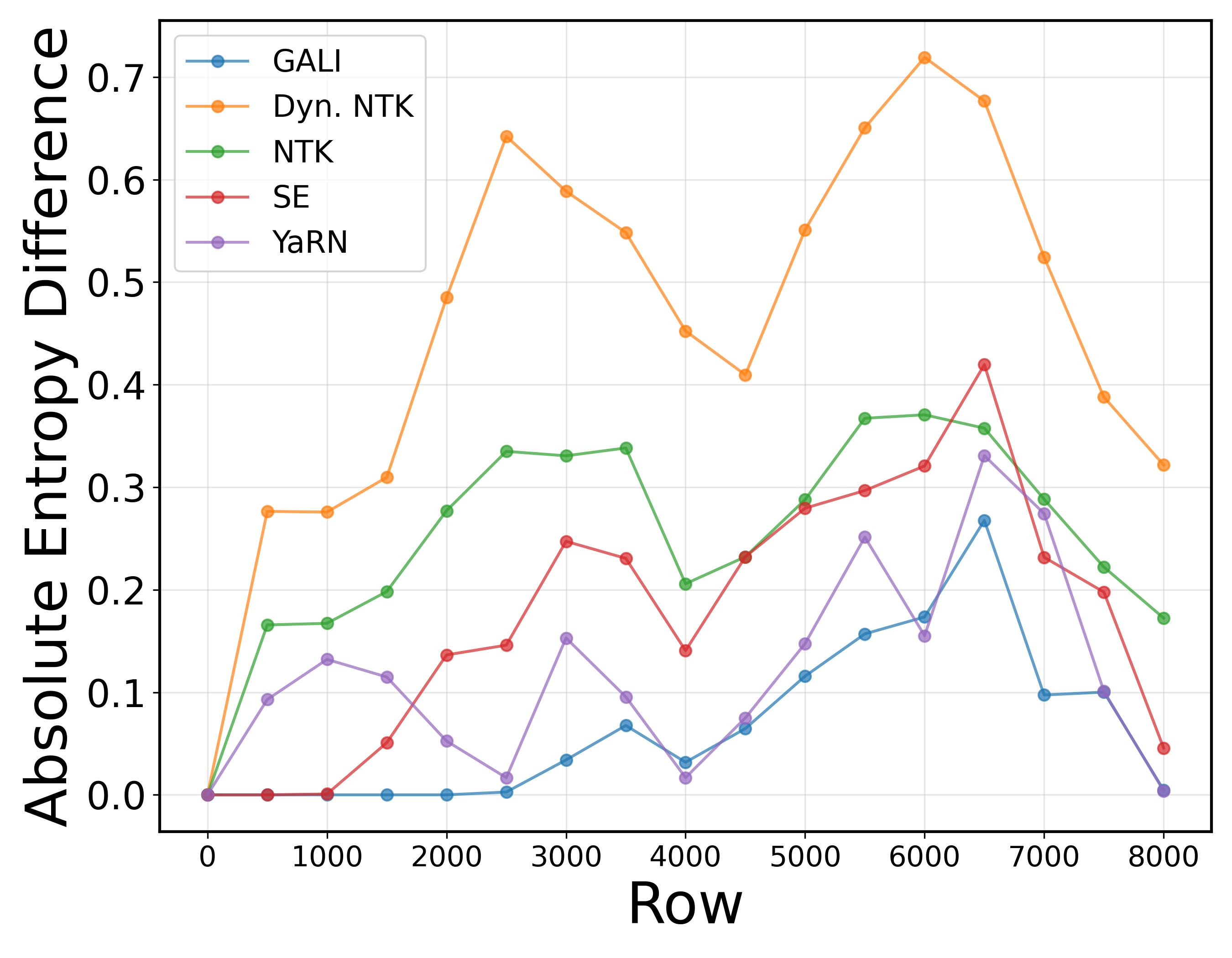}
        \label{fig:attn_row_ent_2k}
    }
    \subfigure[Row Entropy Differences (4K)]{
        \includegraphics[width=0.31\linewidth]{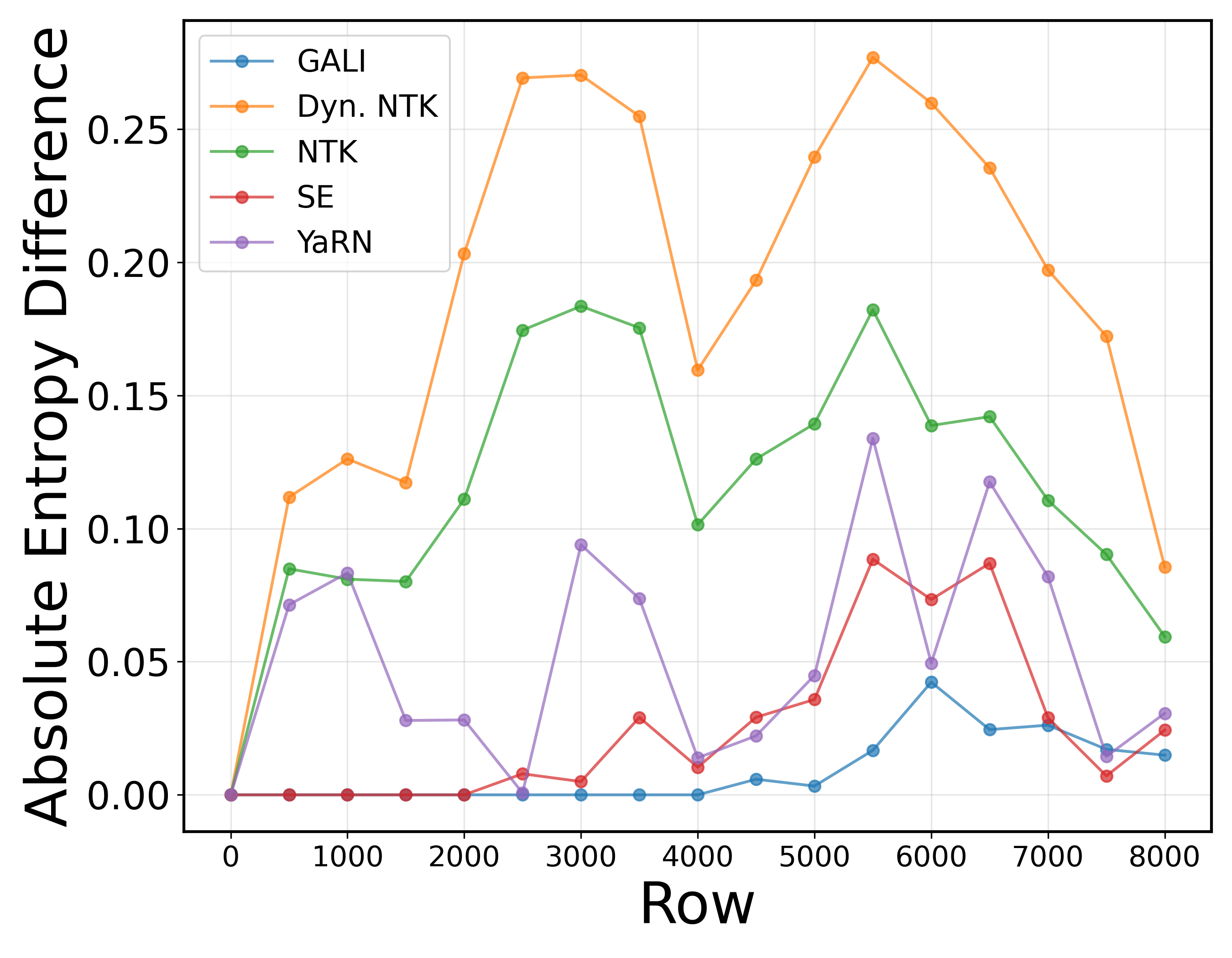}
        \label{fig:attn_row_ent_4k}
    }
    \caption{Differences in attention score metrics and row-wise entropy across methods compared to the original LLM. (a) represents the attention score matrix differences. The values show the sum of the absolute differences between the attention scores of the length interpolation methods and the original model. (b) and (c) show row-wise entropy differences. 2K indicates methods using [0, 2048) positional intervals, while 4k corresponds to [0, 4096). The figures are generated using a sample from NarrativeQA with a prefill length of  8091 tokens. More details on the attention score distributions can be found in Appendix \ref{app:atten_score_matrix}.}
    \label{fig:combined_attn_differences}
\end{figure*}

\subsection{Attention Distribution Analysis}
By analyzing the attention logit distribution of GALI, we observe its local linear interpolation eliminating attention logit outliers produces an interpolated attention distribution that closely matches the original, allowing it to preserve the model’s native behavior when extrapolating to longer inputs. This enables the extrapolation process to fully benefit from the model’s pretrained capabilities.
We design a new experiment to demonstrate the advantage of GALI’s ability to avoid attention logit outliers and maintain an attention distribution that best aligns with the original model. That is, evaluate extrapolation methods while controlling for the model’s inherent positional interval bias. Instead of applying extrapolation across the full training context window, we first restrict each method to a narrower positional interval range. We then extend this range to match the model’s training context window and compare the resulting attention distributions to those produced by the original model. A closer match indicates better alignment with the model’s native positional understanding. Consequently, as the positional understanding of the model improves, the effectiveness of the extrapolation method also improves if its attention distribution remains faithful to the original.

More concretely, we apply various training-free extrapolation methods to Llama3-8b-ins-2k (Llama3-2k) and Llama3-4k, and compare their attention score distributions against that of Llama3-8k. As shown in Figure \ref{fig:attn_differences_grouped}, GALI consistently yields the smallest distribution gap, whether extrapolating from 2k or 4k to 8k. Remarkably, GALI using 2k intervals outperforms Dyn-NTK, NTK, and YaRN using 4k intervals.
In addition to the comparison of the global attention score, we further examine the differences in row-wise attention entropy between Llama3-2k and Llama3-8k, as attention entropy has been shown to strongly correlate with model performance \cite{zhang2024attention, farquhar2024detecting}. As illustrated in Figures \ref{fig:attn_row_ent_2k} and \ref{fig:attn_row_ent_4k}, GALI achieves the smallest row-wise entropy differences among all methods, indicating fewer attention outliers and greater local stability.

In summary, GALI’s attention logit interpolation, eliminating attention logit outliers, preserves the attention score distribution at both global and local levels, which underpins its strong performance in length extrapolation. As backbone models continue to improve their understanding of positional intervals, the ability of GALI to maintain distributional alignment is expected to further enhance its downstream performance.

\begin{figure}[t]
    \centering
    \includegraphics[width=0.8\linewidth]{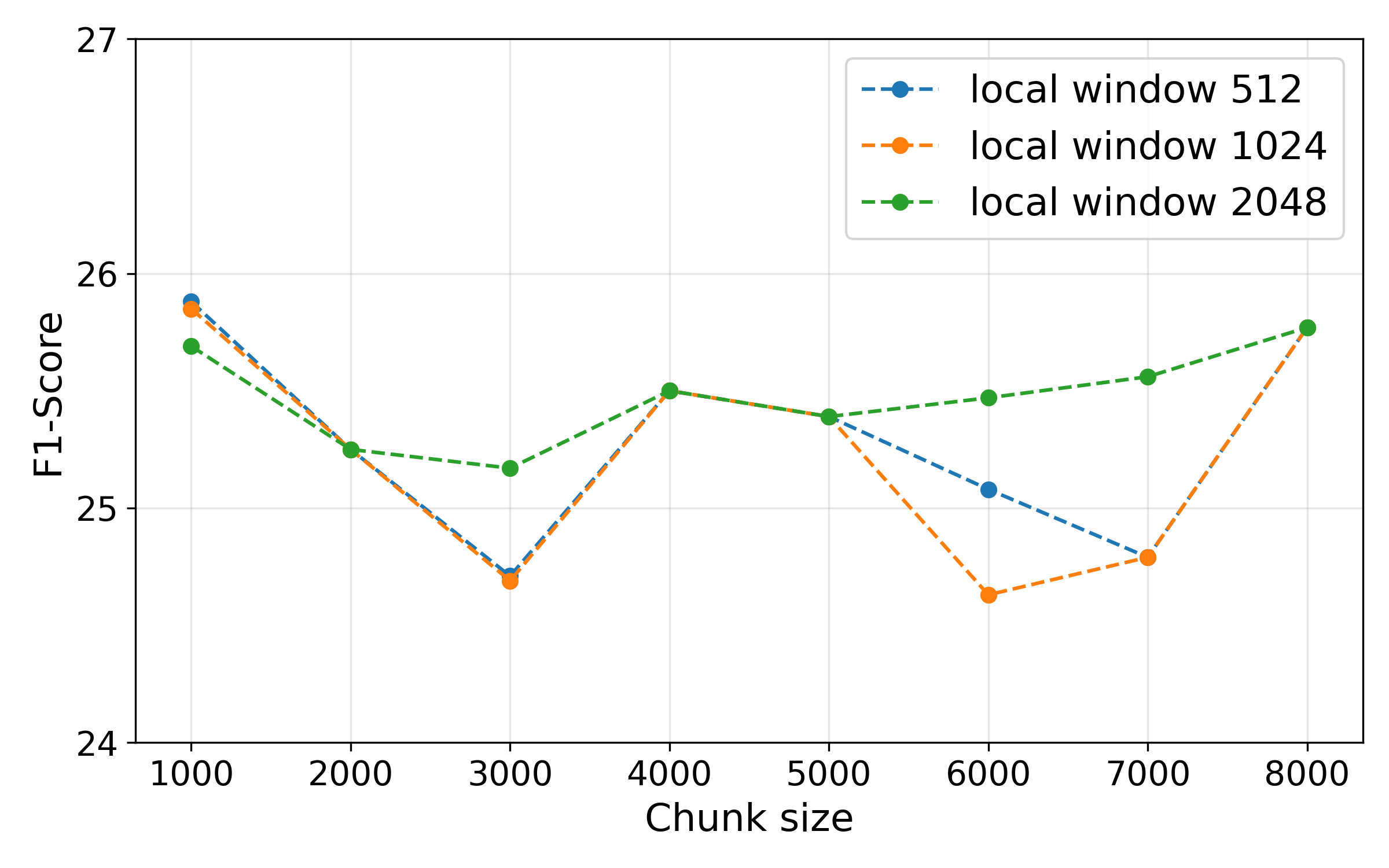}
    \vspace{-0.5cm}
    \caption{Impact of local window and chunk size on performance. The experiments use the Llama3-8b-ins-8k, with an extrapolated context window of 16384 tokens.}
    \label{fig:local_window_group_size}
\end{figure}

\subsection{Ablation Studies}
In this section, we investigate the impact of the size of the local window and the chunk on GALI. We conducted our experiments using NarrativeQA, the longest dataset in LongBench. The results are in Figure \ref{fig:local_window_group_size}. First, we observe that the differences across the three local window sizes are marginal, indicating that attention logit interpolation effectively approximates the true attention score distribution.
Secondly, as the chunk size increases, we hypothesize that the observed effects result from the interplay of two factors. Initially, a smaller size of the chunk aligns better with the design of GALI, which prioritizes leveraging pretrained positional intervals as much as possible while minimizing the number of interpolations for each token. When the chunk size increases, the number of pretrained positional intervals utilized by each token decreases, while the number of interpolated positional intervals increases, leading to performance degradation.
However, as the chunk size grows, more tokens have their positional intervals compressed into a smaller range. As analyzed earlier, performing denser interpolations within a smaller positional interval range, such as [0, 4096), yields better results than performing sparser interpolations over a larger positional interval range, such as [0, 8192). GALI's performance begins to improve.

We also performed an ablation study on the use of Gaussian noise to assess its impact. In Table \ref{tab:noise_ablation_main}, the model exhibited a slight performance degradation when Gaussian noise was removed, suggesting that simulating oscillatory behavior via Gaussian perturbation is indeed beneficial. Nevertheless, due to the overall downward trend of attention logit over long sequences, attention logit interpolation remains effective even in the absence of noise.

\begin{table}[!h]
\fontsize{7}{9}\selectfont
\setlength{\tabcolsep}{5pt}
\centering
\caption{Noise Analysis on LongBench and L-Eval.}

\subfigure[LongBench results]{
\centering
\fontsize{6}{9}\selectfont
\setlength{\tabcolsep}{4pt}
\begin{tabular}{c|ccc}
\specialrule{0.5pt}{0pt}{1pt}
\specialrule{0.5pt}{0pt}{1pt}
\multirow{5}{*}{\rotatebox[origin=c]{90}{\fontsize{6}{9}\selectfont Llama3-8b-ins-8k}} & Model & Noise & Average \\
\cmidrule(lr){2-4}
 & GALI-16k & No  & 44.45 \\
& GALI-16k & Yes & 45.17 \\
& GALI-32k & No  & 44.53 \\
& GALI-32k & Yes & 45.38 \\
\specialrule{0.5pt}{0pt}{1pt}
\end{tabular}
}
\hspace{1pt}
\subfigure[L-Eval results]{
\centering
\fontsize{6}{9}\selectfont
\setlength{\tabcolsep}{4pt}
\begin{tabular}{c|ccc}
\specialrule{0.5pt}{0pt}{1pt}
\specialrule{0.5pt}{0pt}{1pt}
\multirow{5}{*}{\rotatebox[origin=c]{90}{\fontsize{6}{9}\selectfont Llama3-8b-ins-8k}}& Model & Noise & Average \\
\cmidrule(lr){2-4}
 & GALI-16k & No  & 42.01 \\
& GALI-16k & Yes & 42.32 \\
& GALI-32k & No  & 40.16 \\
& GALI-32k & Yes & 42.79 \\
\specialrule{0.5pt}{0pt}{1pt}
\end{tabular}
}

\label{tab:noise_ablation_main}
\end{table}

\section{Conclusion}

The paper introduces Greedy Attention Logit Interpolation (GALI), a training-free method for length extrapolation in LLMs. Our evaluations show GALI achieves stable and superior performance across both short- and long-context tasks without requiring any input-length-specific tuning.
We found that extrapolation within narrower positional ranges can yield better results, even on short context tasks.
GALI avoids computation over position embeddings, making it compatible with other architectures exhibiting long-term decay, such as ALiBi. 

\section*{Limitations}
GALI's current limitation is the need for two passes of attention logit computation, making it incompatible with flash attention. Future work will focus on integrating GALI into flash attention and improving the efficiency of local linear interpolation.



\bibliography{custom}

\clearpage
\appendix
\label{sec:appendix}

\section{Long term decay of RoPE}
\label{app:longtermdecay}
We show another example of the long term decay caused by RoPE in this section.  


\begin{figure}[thp]
    \centering
    \scriptsize
    \subfigure[]{
        \includegraphics[width=0.9\linewidth,trim=10 0 10 5, clip]{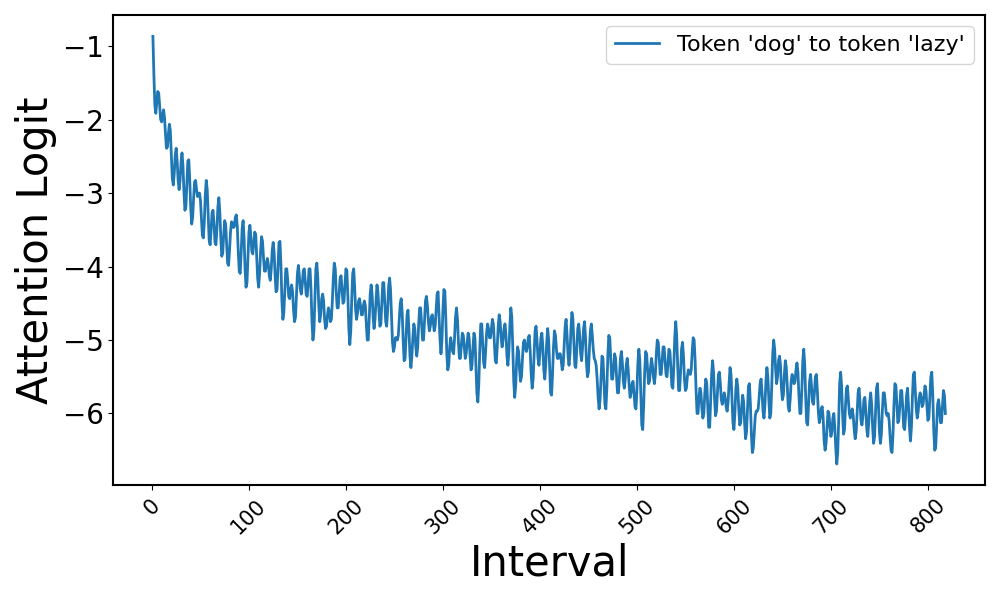}
        \label{fig:logit_-1_-2}
    }
    \subfigure[]{
        \includegraphics[width=0.9\linewidth,trim=10 0 10 5, clip]{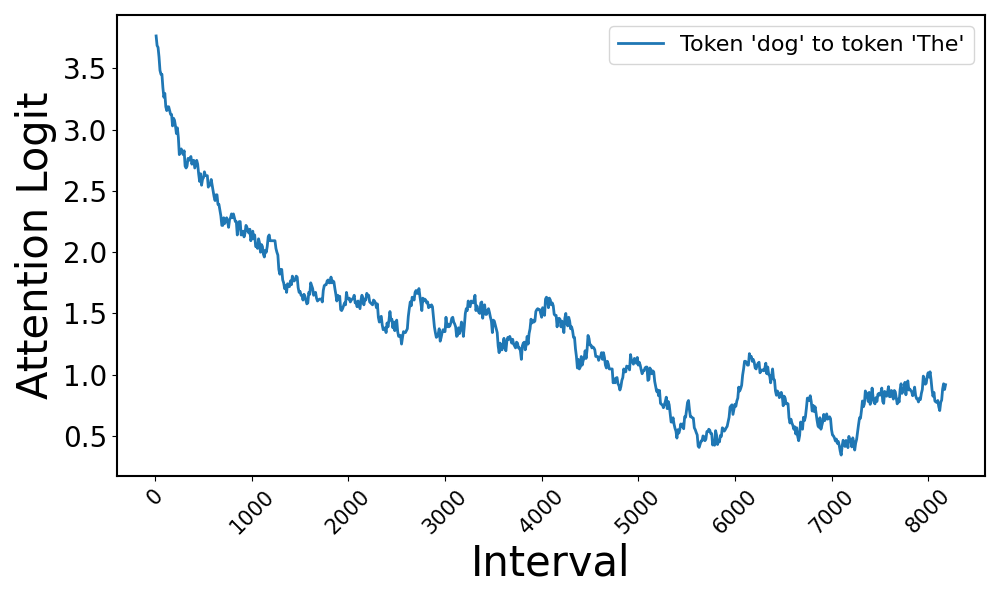}
        \label{fig:logit_-1_0}
    }
    \caption{Visualization of long-term decay in attention logit. The sentence “The quick brown fox jumps over the lazy dog.” is fed into a one-layer Llama3-8b-ins model. Figure \ref{fig:logit_-1_-2} shows the attention logit from the last token to the second-to-last token. Another example can be found in Appendix Figure \ref{fig:logit_-1_0} presents the logit from the last token to the first token. As the positional ID interval increases from 1 to 819, a clear decay phenomenon in the logit is observed.}
    \label{fig:app_logit_long_term_decay}
\end{figure}

\section{Data stastics}
\label{app: data stastics}

\begin{figure*}[htp]
    \centering
    \subfigure[Llama2 Tokenizer]{
        \includegraphics[width=0.9\linewidth]{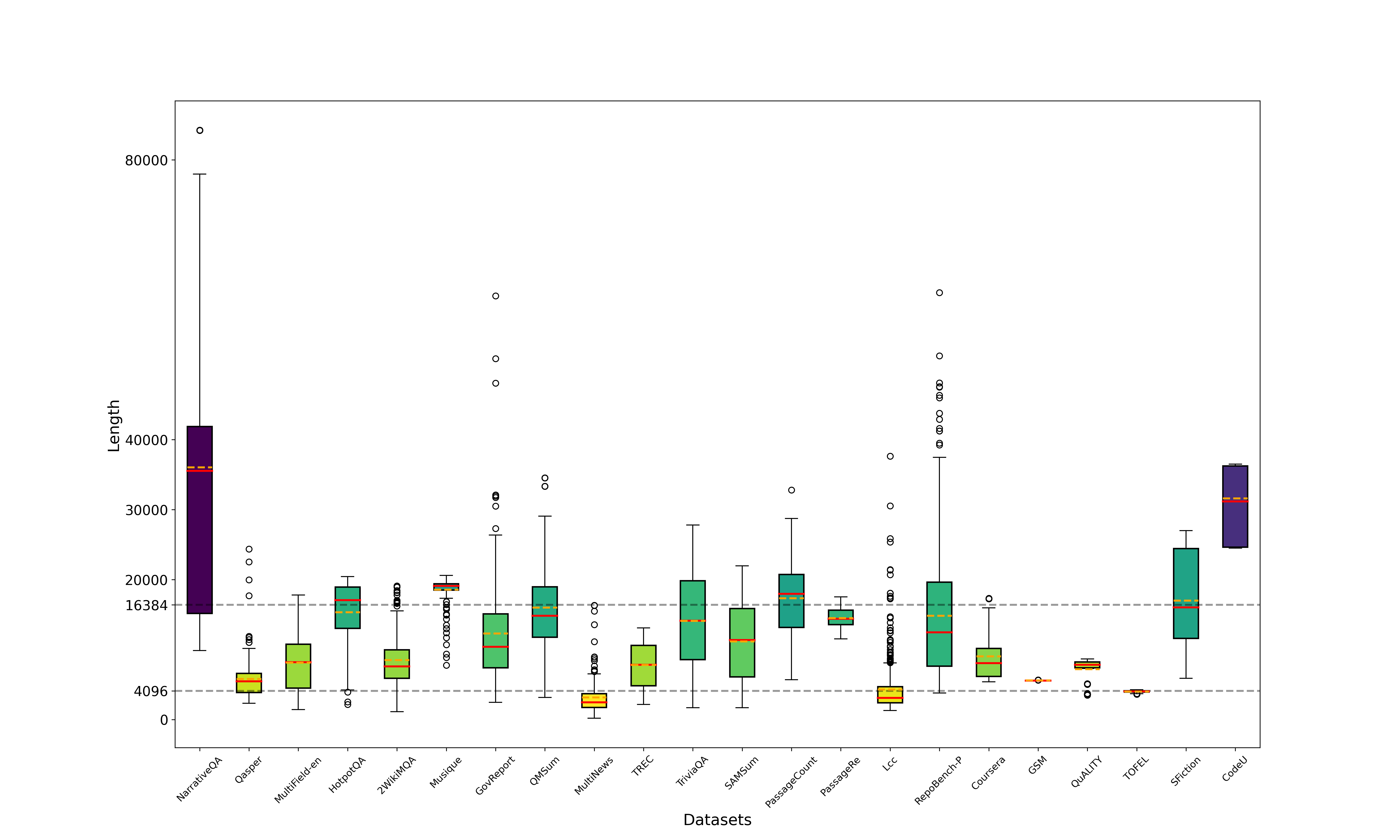}
        \label{fig:llama2_length_dist}
    }
    \subfigure[Llama3 Tokenizer]{
        \includegraphics[width=0.9\linewidth]{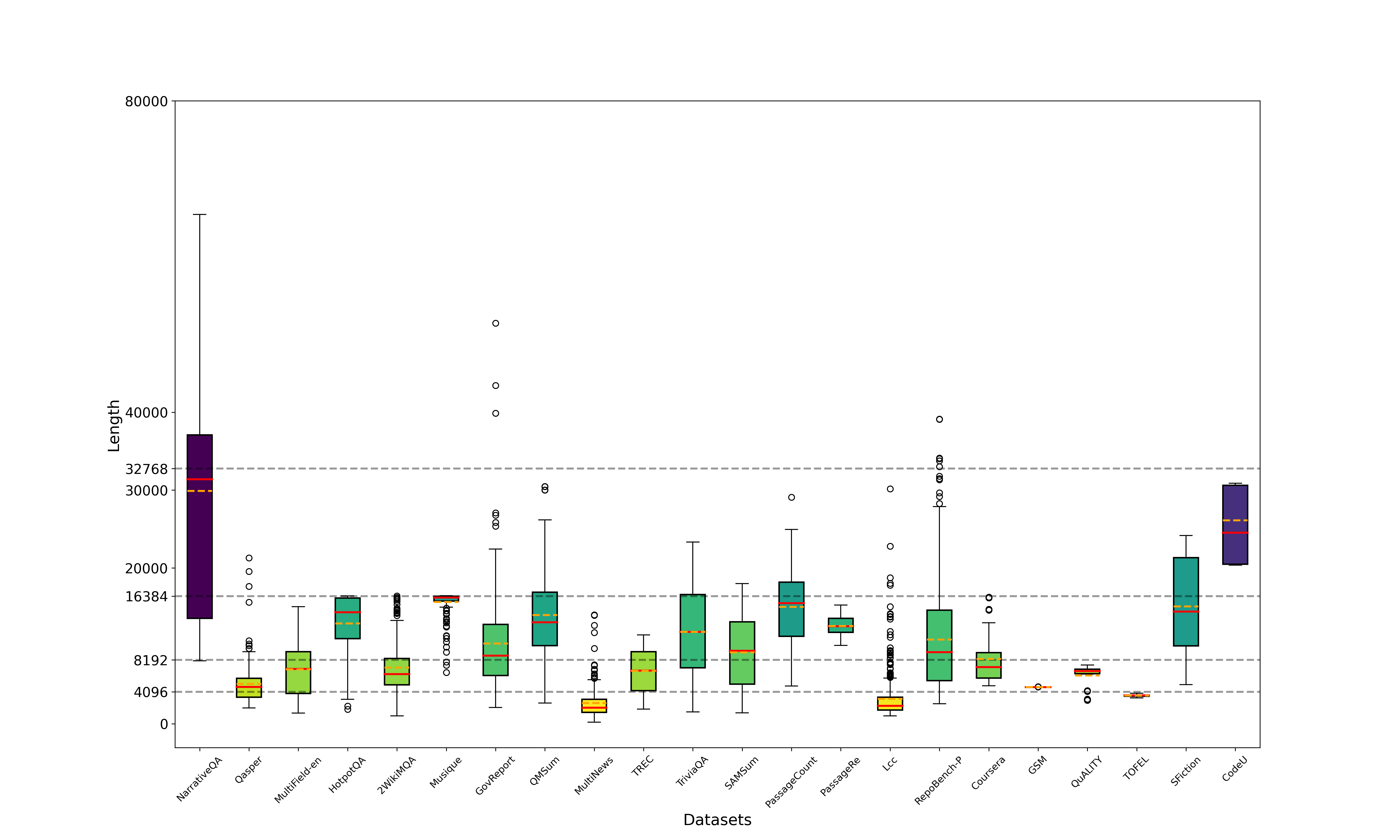}
        \label{fig:llama3_length_dist}
    }
    \caption{Length distributions using Llama2 and Llama3 tokenizers. The left figure shows the distribution with Llama2, and the right figure shows the distribution with Llama3. The red line represents the median, the orange dashed line represents the mean, and the darker the color of the box, the greater the average length.}
    \label{fig:llama_length_dist}
\end{figure*}

In this section, we provide detailed information about each dataset used in LongBench and L-Eval. Table \ref{tab:benchmark_task_type} presents the word length, task type, and number of samples for each dataset. Figure \ref{fig:llama2_length_dist} and \ref{fig:llama3_length_dist} show the length distributions of each dataset using the Llama2 and Llama3 tokenizers, respectively.

\begin{table*}[t]
\centering
\caption{We list the task type, average word lengths, and the number of samples for each dataset we used in our work.} \label{tab:benchmark_task_type}
\begin{tabular}{|l|l|l|r|r|}
    \toprule
    \textbf{Benchmark} & \textbf{Dataset} & \textbf{Task Type} & \textbf{Avg Len} & \textbf{\#Sample} \\
    \hline
    \hline

    \multirow{16}{*}{LongBench} & NarrativeQA & Single-doc QA & 18409 & 200 \\
    & Qasper & Single-doc QA & 3619 & 200 \\
    & MultiField-en & Single-doc QA & 4559 & 150 \\
    & HotpotQA & Multi-doc QA & 9151 & 200 \\
    & 2WikiMQA & Multi-doc QA & 4887 & 200 \\
    & Musique & Multi-doc QA & 11214 & 200 \\
    & GovReport & Summarization & 8734 & 200 \\
    & QMSum & Summarization & 10614 & 200 \\
    & MultiNews & Summarization & 2113 & 200 \\
    & TREC & Few shot & 5177 & 200 \\
    & TriviaQA & Few shot & 8209 & 200 \\
    & SAMSum & Few shot & 6258 & 200 \\
    & PassageCount & Synthetic & 11141 & 200 \\
    & PassageRe & Synthetic & 9289 & 200 \\
    & LCC & Code & 1235 & 500 \\
    & RepoBench-P & Code & 4206 & 500 \\
    \midrule
    \multirow{6}{*}{L-Eval} & Coursera & Multiple choice & 9075 & 172 \\
    & GSM (16-shot) & Solving math problems & 5557 & 100 \\
    & QuALITY & Multiple choice & 7169 & 202 \\
    & TOEFL & Multiple choice & 3907 & 269 \\
    & SFCition & True or False Questions & 16381 & 64 \\
    & CodeU & Deducing program outputs & 31575 & 90 \\
    \bottomrule
\end{tabular}
\end{table*}

\section{Extra experiment results}
\label{app: extra experiment results}

\subsection{Real-world long-context task results}
\label{app:real world task}
We conducted experiments on LongBench and L-Eval using the Llama2-4k backbone, as shown in Tables \ref{tab:longbench_llama2} and \ref{tab:leval_all_llama2}. On LongBench, GALI performed similarly to NTK, Dyn-NTK, and YaRN, but was weaker than SelfExtend and ChunkLlama. However, all methods performed significantly worse than those using the Llama3-8b-ins-4k backbone.

Although Llama2-7b-chat and Llama3-8b-ins have similar parameter scales, Llama3 demonstrates a deeper understanding of pretrained positional intervals closer to its training context window. Consequently, GALI performed significantly better on Llama3-8b-ins-4k than on Llama2-4k, with similar trends observed across other methods. As the quality of the pretrained model improves, it better aligns with GALI’s principle of maximizing the use of pretrained positional intervals.

Regarding the best-performing method on Llama2-4k, SelfExtend has been reported to be highly sensitive to hyperparameters \cite{Jin2024LLMML}. Specifically, larger group sizes and smaller local windows sometimes yield better results, which supports our conclusion in Section \ref{main:Real-world long-context task results}. These configurations emphasize the use of smaller positional intervals, reducing reliance on larger ones and preventing content from being placed in less well-understood positional intervals. This limitation affects GALI’s effectiveness on Llama2, as GALI assumes the model fully understands its entire training context window, thereby always maximizing the use of pretrained positional intervals.

On the L-Eval benchmark, the performance gap between GALI and the best approaches was smaller than on LongBench. This is because, when using the Llama2 tokenizer, datasets such as Coursera, GSM, QuALITY, and TOEFL in L-Eval are much shorter than 16k, allowing all methods to leverage Llama2-4k’s well-understood smaller relative positional intervals. In longer datasets like SFictions and CodeU, performance is task-dependent. SFictions is a True/False task with higher results, while CodeU is a code inference task with much lower results. We also report complete results using the Llama3-8k backbone. Our method performed almost identically to the backbone model, as the token lengths of GSM, QuALITY, and TOEFL are all below 8192. However, SelfExtend, NTK, and YaRN outperformed the backbone model, further validating our conclusion that even on short text datasets, using a smaller range of positional intervals for length extrapolation leads to better task performance.

\begin{table*}[!htp]
\fontsize{18}{24}\selectfont
\setlength{\tabcolsep}{5pt}
\centering
\caption{Performance comparison with different backbone LLMs and training-free length extrapolation methods. The best result in each experiment has been bolded. 
{\color{green}\textsuperscript{*}} indicates the results reported by LongBench\cite{bai-etal-2024-longbench}, 
{\color{blue}\textsuperscript{*}} indicates the results reported by LongBench\cite{Jin2024LLMML}. 
The number following each method represents the target context window. For example, 16k means 16 × 1024. The "Original" means testing with the backbone model, i.e., the model shown in the left column. 
}
\label{tab:longbench_llama2}
\resizebox{\textwidth}{!}{
\begin{tabular}{c|cccccccccccccccccc}
\specialrule{1pt}{0pt}{2pt}
\toprule
 &\multirow{4}{*}{~~~~~~~~~~~~~\textbf{Methods}~~~~~~~~~~~~~} & \multicolumn{3}{c}{\textbf{Single document QA}} & \multicolumn{3}{c}{\textbf{Multi document QA}} & \multicolumn{3}{c}{\textbf{Summarization}} & \multicolumn{3}{c}{\textbf{Few-shot Learning}} & \multicolumn{2}{c}{\textbf{Synthetic}} & \multicolumn{2}{c}{\textbf{Code}} & \multirow{4}{*}{\textbf{Average}} \\
\cmidrule(r){3-5} \cmidrule(r){6-8} \cmidrule(r){9-11} \cmidrule(r){12-14} \cmidrule(r){15-16} \cmidrule(r){17-18}
& & \rotatebox{30}{NarrativeQA} & \rotatebox{30}{Qasper} & \rotatebox{30}{MultiField-en} & \rotatebox{30}{HotpotQA} & \rotatebox{30}{2WikiMQA} & \rotatebox{30}{Musique} & \rotatebox{30}{GovReport} & \rotatebox{30}{QMSum} & \rotatebox{30}{MultiNews} & \rotatebox{30}{TREC} & \rotatebox{30}{TriviaQA} & \rotatebox{30}{SAMSum} & \rotatebox{30}{PassageCount} & \rotatebox{30}{PassageRe} & \rotatebox{30}{Lcc} & \rotatebox{30}{RepoBench-P} & \\

\midrule

\multirow{8}{*}{\rotatebox[origin=c]{90}{\fontsize{18}{100}\selectfont Llama2-7b-chat-4k}}  &  Original\color{green}\textsuperscript{*} & 18.70 & 19.20 & 36.80 & 25.40 & 32.80 & 9.40 & 27.30 & 20.80 & 25.80 & 61.50 & 77.80 & 40.70 & 2.10 & 9.80 & 52.40 & 43.80 & \cellcolor{YellowGreen!25}31.52 \\
 &  Original  &  8.48  &  13.97  &  20.4  &  13.62  &  16.77  &  5.46  &  25.17  &  12.47  &  24.78  &  67.5  &  74.24  &  40.28  &  2.30  &  3.25  &  56.39  &  50.36  &  \cellcolor{YellowGreen!25}27.22 \\
 &  SelfExtend-16k\color{blue}\textsuperscript{*} &  21.69  & 25.02 & 35.21 & 34.34 & 30.24 & 14.13 & 27.32 & 21.35 & 25.78 & 69.50 & 81.99 & 40.96 & 5.66 & 5.83 & 60.60 & 54.33 & \cellcolor{YellowGreen!25}\textbf{34.62} \\
 &  SelfExtend-16k & 6.89 & 12.67 & 25.95 & 9.08 & 11.25 & 5.88 & 26.80 & 16.39 & 22.79 & 67.50 & 69.88 & 41.18 & 2.18 & 3.21 & 58.21 & 51.65 & \cellcolor{YellowGreen!25}26.97 \\
 &  ChunkLlama-16k & 8.48 & 13.97 & 20.40 & 13.62 & 16.77 & 5.46 & 25.17 & 12.47 & 24.78 & 67.50 & 74.24 & 40.28 & 2.30 & 3.25 & 56.39 & 50.36 & \cellcolor{YellowGreen!25}27.22 \\
 &  NTK-16k & 0.73 & 10.33 & 19.44 & 2.38 & 7.91 & 0.42 & 19.47 & 6.26 & 26.13 & 59.50 & 17.89 & 23.17 & 0.52 & 0.51 & 50.70 & 27.91 & \cellcolor{YellowGreen!25}17.08 \\
 &  Dyn-NTK-16k  & 3.79 & 10.37 & 22.38 & 7.47 & 10.26 & 3.81 & 29.52 & 20.13 & 22.84 & 63.50 & 45.35 & 31.79 & 2.29 & 4.33 & 57.13 & 42.16 & \cellcolor{YellowGreen!25}23.57 \\
 &  YaRN-16k & 3.22 & 10.86 & 22.14 & 5.52 & 13.36 & 1.32 & 24.78 & 10.90 & 25.92 & 64.50 & 40.60 & 32.36 & 2.20 & 2.15 & 51.74 & 43.91 & \cellcolor{YellowGreen!25}22.22 \\
 & \cellcolor{WildStrawberry!25}\textbf{(Ours)GALI-16k} & \cellcolor{WildStrawberry!25}6.29 & \cellcolor{WildStrawberry!25}16.73 & \cellcolor{WildStrawberry!25}22.26 & \cellcolor{WildStrawberry!25}12.82 & \cellcolor{WildStrawberry!25}13.65 & \cellcolor{WildStrawberry!25}6.31 & \cellcolor{WildStrawberry!25}23.58 & \cellcolor{WildStrawberry!25}15.96 & \cellcolor{WildStrawberry!25}23.37 & \cellcolor{WildStrawberry!25}62.00 & \cellcolor{WildStrawberry!25}72.80 & \cellcolor{WildStrawberry!25}25.12 & \cellcolor{WildStrawberry!25}1.83 & \cellcolor{WildStrawberry!25}2.83 & \cellcolor{WildStrawberry!25}58.71 & \cellcolor{WildStrawberry!25}48.51 & \cellcolor{YellowGreen!25}25.80 \\


\bottomrule
\end{tabular}
}
\end{table*}

\begin{table}[!t]
\fontsize{18}{24}\selectfont
\setlength{\tabcolsep}{5pt}
\centering
\caption{Performance comparison with different backbone LLMs and training-free length extrapolation methods. The best result in each experiment has been bolded.
{\color{orange}\textsuperscript{*}} indicates the results reported by ChunkLlama\cite{an2024training}, {\color{blue}\textsuperscript{*}} indicates the results reported by LongBench\cite{Jin2024LLMML}, and {\color{red}\textsuperscript{*}} indicates the results reported by L-Eval\cite{an-etal-2024-l}.}

\label{tab:leval_all_llama2}
\resizebox{0.49\textwidth}{!}{
\begin{tabular}{c|cccccccc}
\specialrule{1pt}{0pt}{2pt}
\toprule
 & ~Methods~ & Coursera & GSM & QuALITY& TOFEL & SFiction & CodeU & Average \\
\midrule

\multirow{8}{*}{\rotatebox[origin=c]{90}{\fontsize{18}{100}\selectfont Llama2-7b-chat-4k}} & 
Original\color{red}{\textsuperscript{*}} & 29.21 & 19.00 & 37.62 & 51.67 & 60.15 & 1.11 & \cellcolor{YellowGreen!25}33.12 \\
& Original & 29.80 & 29.00 & 37.62 & 58.36 & 60.16 & 1.11 & \cellcolor{YellowGreen!25}36.01\\
& SelfExtend-16k\color{blue}{\textsuperscript{*}} & \textbf{35.76} & 25.00 & 41.09 & 55.39 & 57.81 & 1.11 & \cellcolor{YellowGreen!25}36.02 \\
& SelfExtend-16k & 32.99 & 29.00 & 40.59 & 57.62 & 57.81 & 2.22 & \cellcolor{YellowGreen!25}36.71 \\
& ChunkLlama\color{orange}\textsuperscript{*} & 32.12 & \textbf{31.00} & 35.14 & 57.62 & 61.72 & 2.22 & \cellcolor{YellowGreen!25}36.64 \\
& ChunkLlama-16k & 28.92 & \textbf{31.00} & \textbf{43.07} & 58.36 & 60.94 & 2.22 & \cellcolor{YellowGreen!25}\textbf{37.42} \\
& NTK-16k\color{red}{\textsuperscript{*}} & 32.71 & 19.00 & 33.16 & 52.78 & \textbf{64.84} & 0.00 & \cellcolor{YellowGreen!25}33.75 \\
& NTK-16k & 26.89 & 16.00 & 33.66 & \textbf{60.97} & 41.41 & 0.00 & \cellcolor{YellowGreen!25}29.82 \\
& Dyn-NTK\color{orange}\textsuperscript{*} & 13.95 & 13.00 & 30.69 & 52.27 & 57.02 & 1.11 & \cellcolor{YellowGreen!25}28.01 \\
& Dyn-NTK-16k & 15.41 & 13.00 & 33.17 & 54.65 & 54.69 & 1.11 & \cellcolor{YellowGreen!25}28.67 \\
& YaRN-16k & 36.49 & 18.00 & 42.08 & 57.62 & 42.97 & \textbf{7.78} & \cellcolor{YellowGreen!25}34.15  \\
& \cellcolor{WildStrawberry!25}\textbf{(Ours)GALI-16k} & \cellcolor{WildStrawberry!25}35.32 & \cellcolor{WildStrawberry!25}29.00 & \cellcolor{WildStrawberry!25}39.11 & \cellcolor{WildStrawberry!25}54.65 & \cellcolor{WildStrawberry!25}51.43 & \cellcolor{WildStrawberry!25}4.44 & \cellcolor{YellowGreen!25}35.66 \\

\midrule

\multirow{8}{*}{\rotatebox[origin=c]{90}{\fontsize{18}{100}\selectfont Llama3-8b-ins-8k}} & 
Original & 53.05 & 58.00 & 61.88 & 82.16 & 60.16 & 4.44 & \cellcolor{YellowGreen!25}53.93 \\
& SelfExtend-16k & 55.38 & 63.00 & 62.87 & 82.16 & 64.06 & 5.56 & \cellcolor{YellowGreen!25}55.76 \\
& ChunkLlama\color{orange}\textsuperscript{*} & \textbf{56.24} & 54.00 & \textbf{63.86} & 83.27 & \textbf{70.31} & 5.56 & \cellcolor{YellowGreen!25}55.54 \\
& ChunkLlama-16k & 53.34 & 54.00 & 60.89 & 81.78 & 61.72 & 5.56 & \cellcolor{YellowGreen!25}53.53 \\
& NTK-16k & 52.03 & \textbf{77.00} & 65.35 & 81.04 & 42.97 & 0.00 & \cellcolor{YellowGreen!25}56.58 \\
& Dyn-NTK-16k & 52.03 & 55.00 & 61.88 & 82.16 & 52.34 & 2.22 & \cellcolor{YellowGreen!25}52.89 \\
& YaRN-16k & 55.96 & 75.00 & 63.37 & 79.93 & 62.50 & 5.56 & \cellcolor{YellowGreen!25}\textbf{57.83} \\
& \cellcolor{WildStrawberry!25}\textbf{(Ours)GALI-16k} & \cellcolor{WildStrawberry!25}54.65 & \cellcolor{WildStrawberry!25}59.09 & \cellcolor{WildStrawberry!25}61.88 & \cellcolor{WildStrawberry!25}\textbf{83.33} & \cellcolor{WildStrawberry!25}65.63 & \cellcolor{WildStrawberry!25}\textbf{6.67} & \cellcolor{YellowGreen!25}55.42 \\

\midrule

\multirow{6}{*}{\rotatebox[origin=c]{90}{\fontsize{18}{100}\selectfont Llama3-8b-ins-8k}} & SelfExtend-32k & 53.92 & 77.00 & 63.37 & 79.93 & 65.63 & 3.33 & \cellcolor{YellowGreen!25}57.20 \\
& ChunkLlama-32k & 54.36 & 55.00 & 60.89 & 81.78 & 64.06 & 5.56 & \cellcolor{YellowGreen!25}53.61 \\
& NTK-32k & \textbf{58.28} & \textbf{83.00} & \textbf{63.86} & 81.04 & 59.38 & 1.11 & \cellcolor{YellowGreen!25}57.78 \\
& Dyn-NTK-32k & 54.36 & 55.00 & 61.88 & 82.16 & 64.06 & 6.67 & \cellcolor{YellowGreen!25}54.02 \\
& YaRN-32k & 55.23 & 76.00 & 62.38 & 79.18 & \textbf{67.19} & \textbf{7.78} & \cellcolor{YellowGreen!25}\textbf{57.96} \\
& \cellcolor{WildStrawberry!25}\textbf{(Ours)GALI-32k} & \cellcolor{WildStrawberry!25}54.17 & \cellcolor{WildStrawberry!25}59.09 & \cellcolor{WildStrawberry!25}62.38 & \cellcolor{WildStrawberry!25}\textbf{82.68} & \cellcolor{WildStrawberry!25}66.41 & \cellcolor{WildStrawberry!25}\textbf{7.78} & \cellcolor{YellowGreen!25}55.29 \\

\bottomrule
\end{tabular}
}
\end{table}

\subsection{Long language modeling task results}
\label{app:ppl_test}
We also conducted PPL evaluations on the Llama2-7b-chat-4k backbone. As shown in Table \ref{tab:pg19_models_llama2_4k}, GALI maintained a stable PPL, while Dyn-NTK consistently produced the worst results.

\begin{table}[!h]
\fontsize{18}{24}\selectfont
\setlength{\tabcolsep}{5pt}
\centering
\caption{Performance of various methods on PG19 Dataset with different context windows, using Llama2-7b-chat-4k as the backbone model.}
\label{tab:pg19_models_llama2_4k}
\resizebox{0.49\textwidth}{!}{
\begin{tabular}{c|ccccccccc}
\toprule
& \textbf{Methods} & \textbf{1k} & \textbf{2k} & \textbf{3k} & \textbf{4k} & \textbf{5k} & \textbf{6k} & \textbf{7k} & \textbf{8k} \\ \midrule
\multirow{6}{*}{\rotatebox[origin=c]{90}{\fontsize{18}{100}\selectfont Llama3-7b-chat-4k}} & SelfExtend & 8.81 & 8.99 & 9.16 & 9.24 & 9.25 & 9.16 & 9.2 & 9.3 \\
& ChunkLlama & 9.07 & 9.26 & 9.41 & 9.45 & 9.43 & 9.31 & 9.31 & 9.39 \\
& NTK  & 8.95 & 9.04 & 9.16 & 9.18 & 9.16 & 9.06 & 9.26 & 13.67 \\
& Dyn-NTK  & 8.81 & 8.99 & 9.15 & 10.79 & 44.32 & 87.35 & 160.07 & 224.07 \\
& YaRN  & 11.65 & 8.97 & 9.07 & 9.16 & 9.17 & 9.15 & 9.03 & 9.04 \\
& \cellcolor{WildStrawberry!25}\textbf{(Ours)GALI} & \cellcolor{WildStrawberry!25}8.81 & \cellcolor{WildStrawberry!25}8.99 & \cellcolor{WildStrawberry!25}9.15 & \cellcolor{WildStrawberry!25}9.24 & \cellcolor{WildStrawberry!25}9.59 & \cellcolor{WildStrawberry!25}9.66 & \cellcolor{WildStrawberry!25}9.63 & \cellcolor{WildStrawberry!25}9.66 \\ 
\bottomrule
\end{tabular}
}
\end{table}

\subsection{Attention distribution analysis results}
\label{app:atten_score_matrix}
\begin{figure*}[htp]
    \centering
    \subfigure[Llama3-8b-ins-2k backbone]{
        \includegraphics[width=0.9\linewidth]{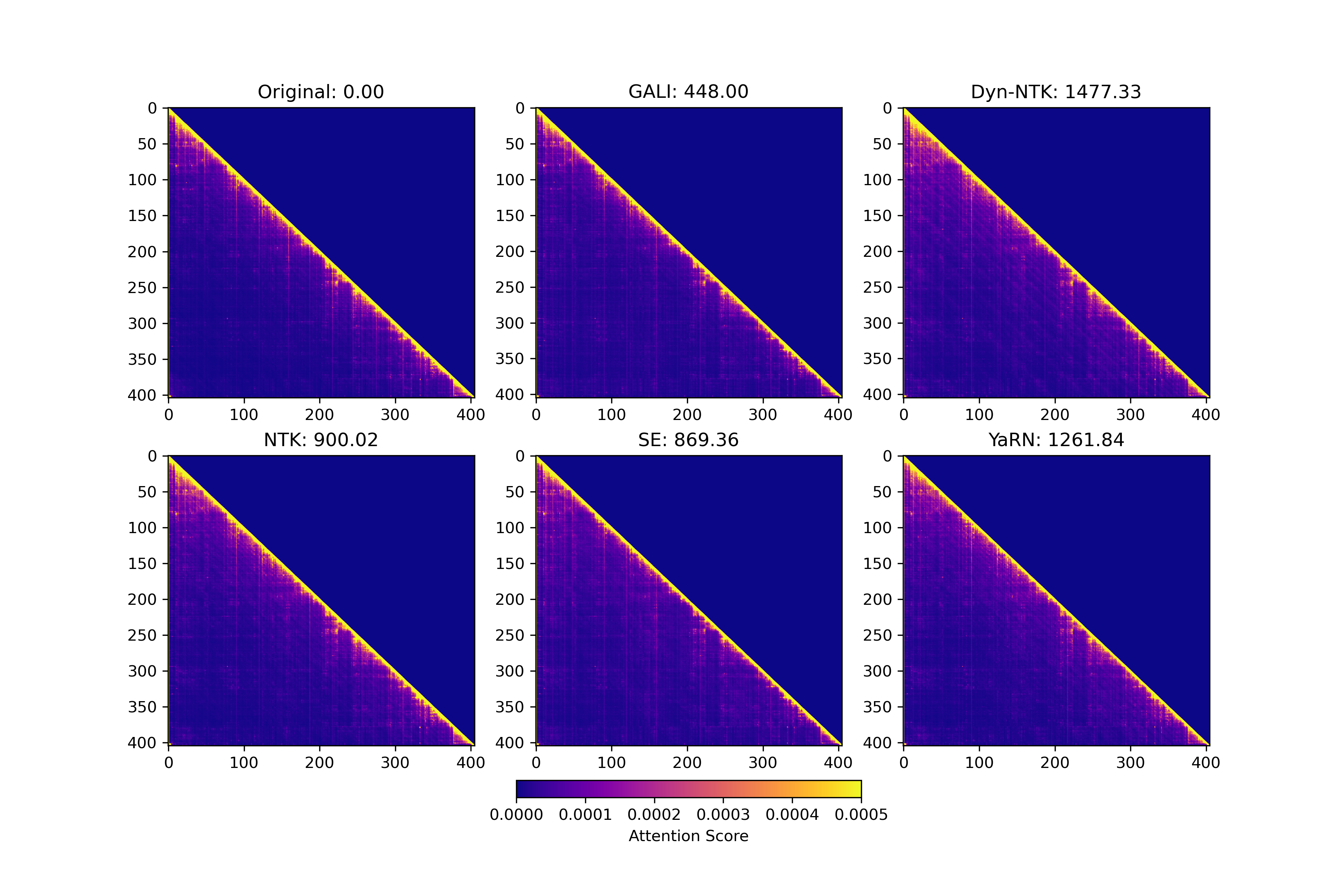}
        \label{fig:attn_2k_dist}
    }
    \subfigure[Llama3-8b-ins-4k backbone]{
        \includegraphics[width=0.9\linewidth]{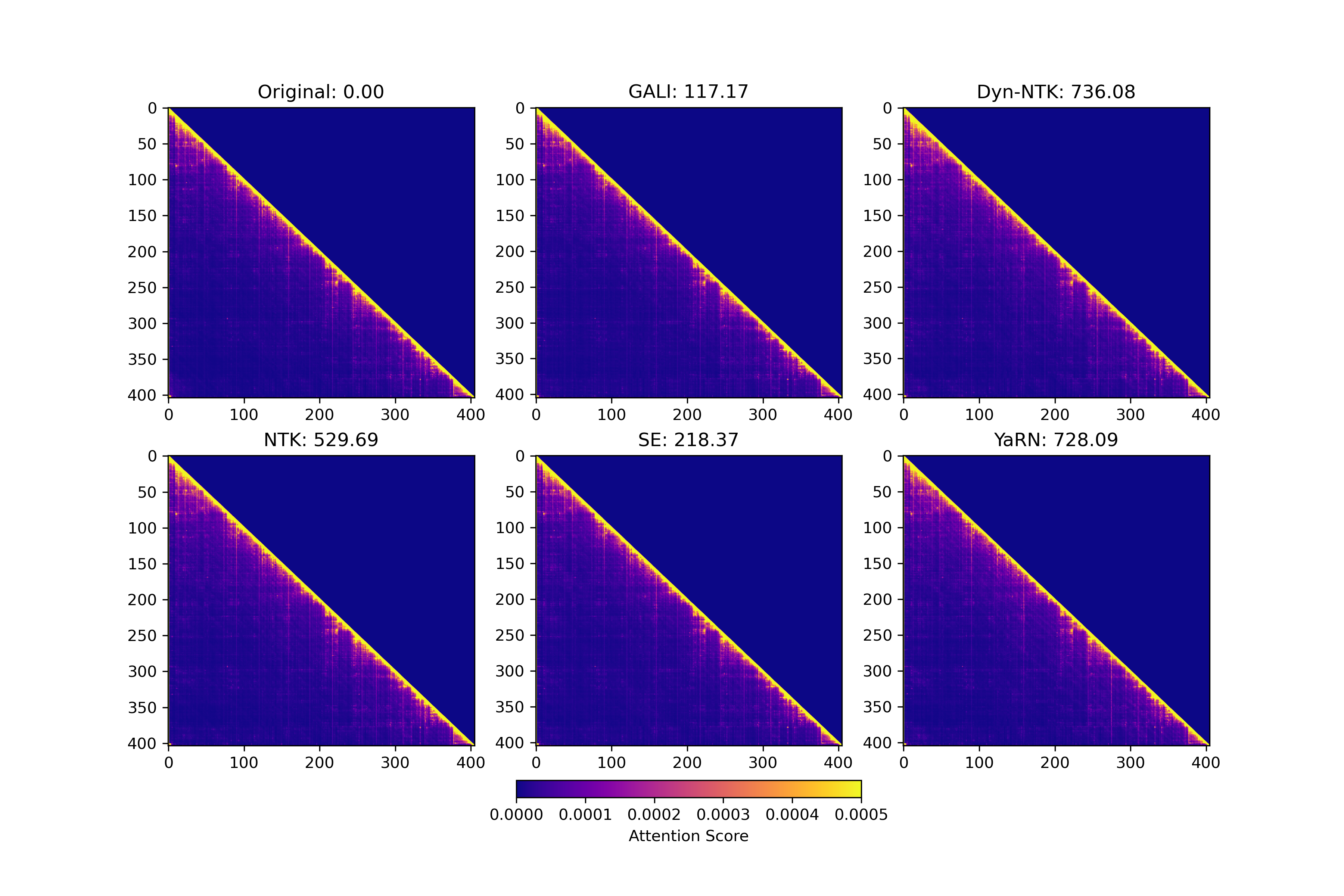}
        \label{fig:attn_4k_dist}
    }
    \caption{This is a comparison of the attention score matrices obtained using Llama3-2k and Llama3-4k for length extrapolation with those of Llama3-8k. Note that we averaged the attention scores across all layers and heads, applied average pooling to scale the matrix to 0.05\%, and set the maximum value of the heatmap to 0.0005 for better visualization. “Original” represents the attention score matrix of Llama3-8k, and the number next to each method’s name indicates the sum of the absolute differences between the method’s attention score matrix and the “Original” matrix.}
    \label{fig:attn_score_dist}
\end{figure*}

In this section, we present the detailed results of the attention distribution analysis. First, we compare the differences between the attention score matrix of length extrapolation methods and the standard attention score matrix, as shown in Figure \ref{fig:attn_score_dist}. For this analysis, we averaged the attention score matrices for each layer and each head before comparison. Whether comparing Llama3-2k or Llama3-4k, GALI consistently achieved the highest similarity to the standard attention score matrix. Additionally, we observed that all methods exhibited higher values in the lower-left corner of the matrix compared to the standard attention score matrix. We attribute this to the fact that these methods do not perform true extrapolation, whereas the standard Llama3-8k model, utilizing a larger positional interval range [0, 8192), results in a lower mean value of the attention scores.

We also analyzed the attention score distribution by extracting 8 rows from the attention score matrix, with the results shown in Figures 5 and 6. The figures clearly demonstrate that GALI’s attention score distribution for each row is closer to the corresponding original attention score distribution. Moreover, as the row index increases, the attention score distributions of all length extrapolation methods show an upward shift relative to the original attention score distribution. This aligns with our earlier analysis, as using a smaller positional interval range results in higher mean value of the attention scores.

\begin{figure*}[thp]
    \centering
    \subfigure[Attention scores of row 1000]{
        \includegraphics[width=0.45\linewidth]{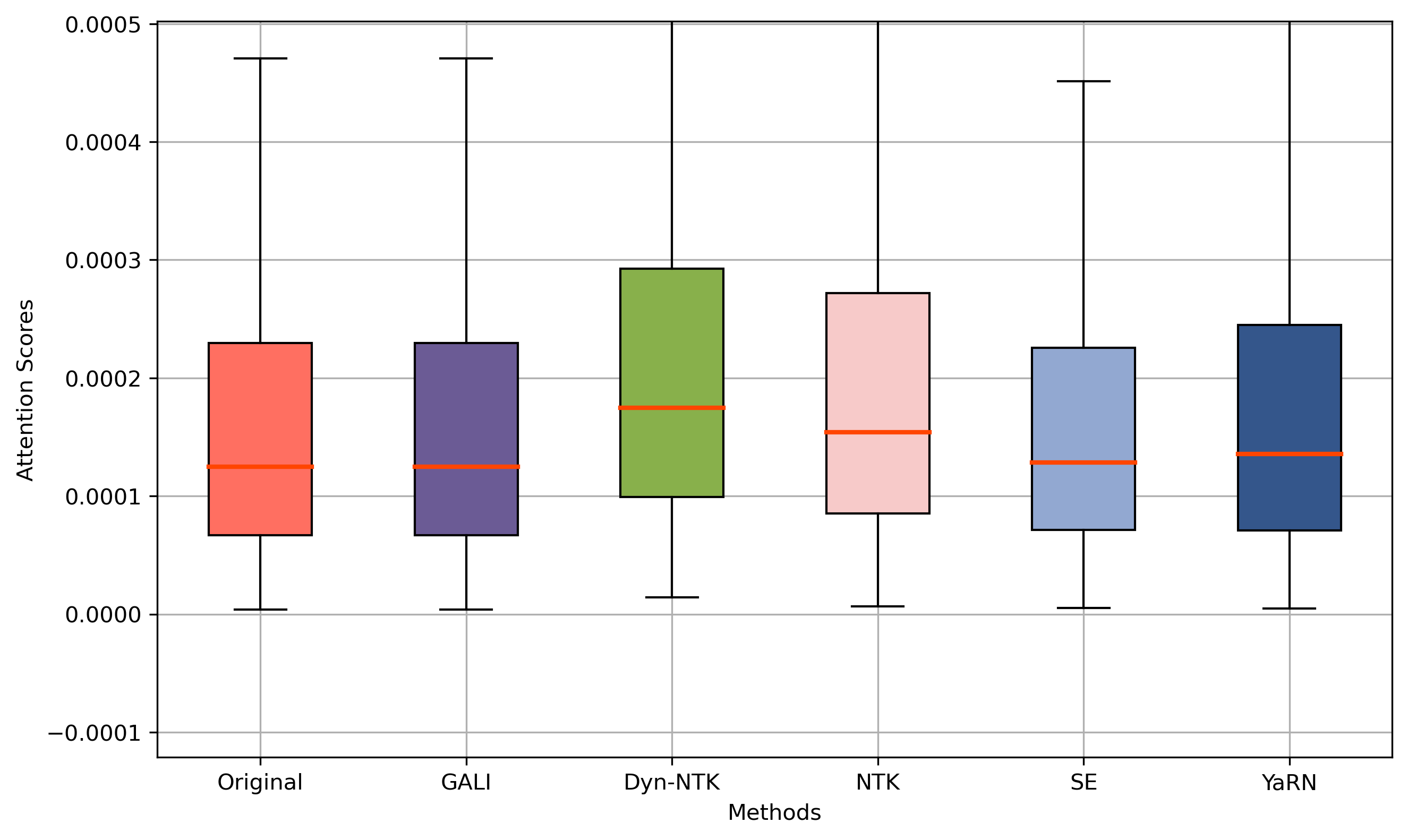}
        \label{fig:box_2k_1000}
    }
    \subfigure[Attention scores of row 2000]{
        \includegraphics[width=0.45\linewidth]{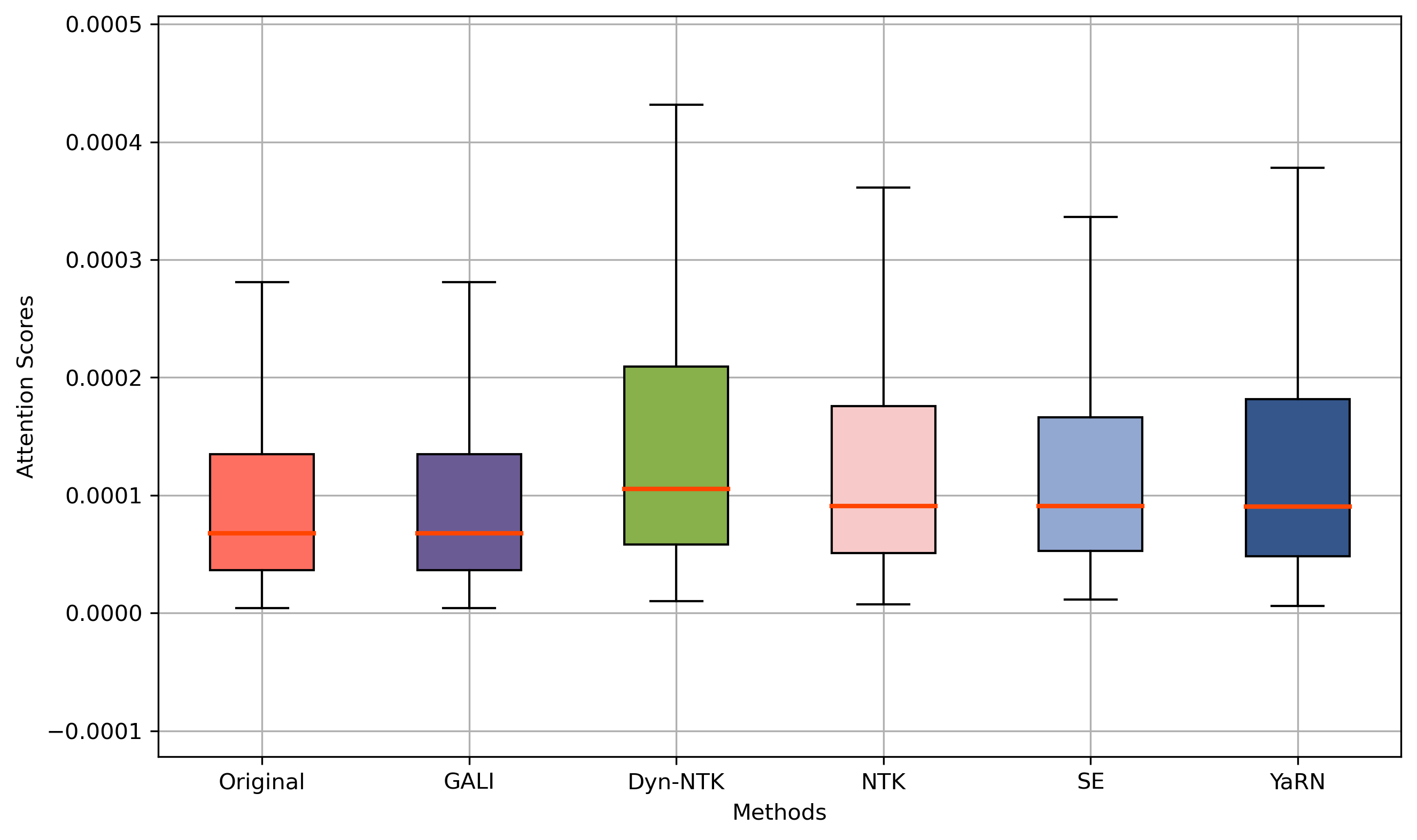}
        \label{fig:box_2k_2000}
    }
    \subfigure[Attention scores of row 3000]{
        \includegraphics[width=0.45\linewidth]{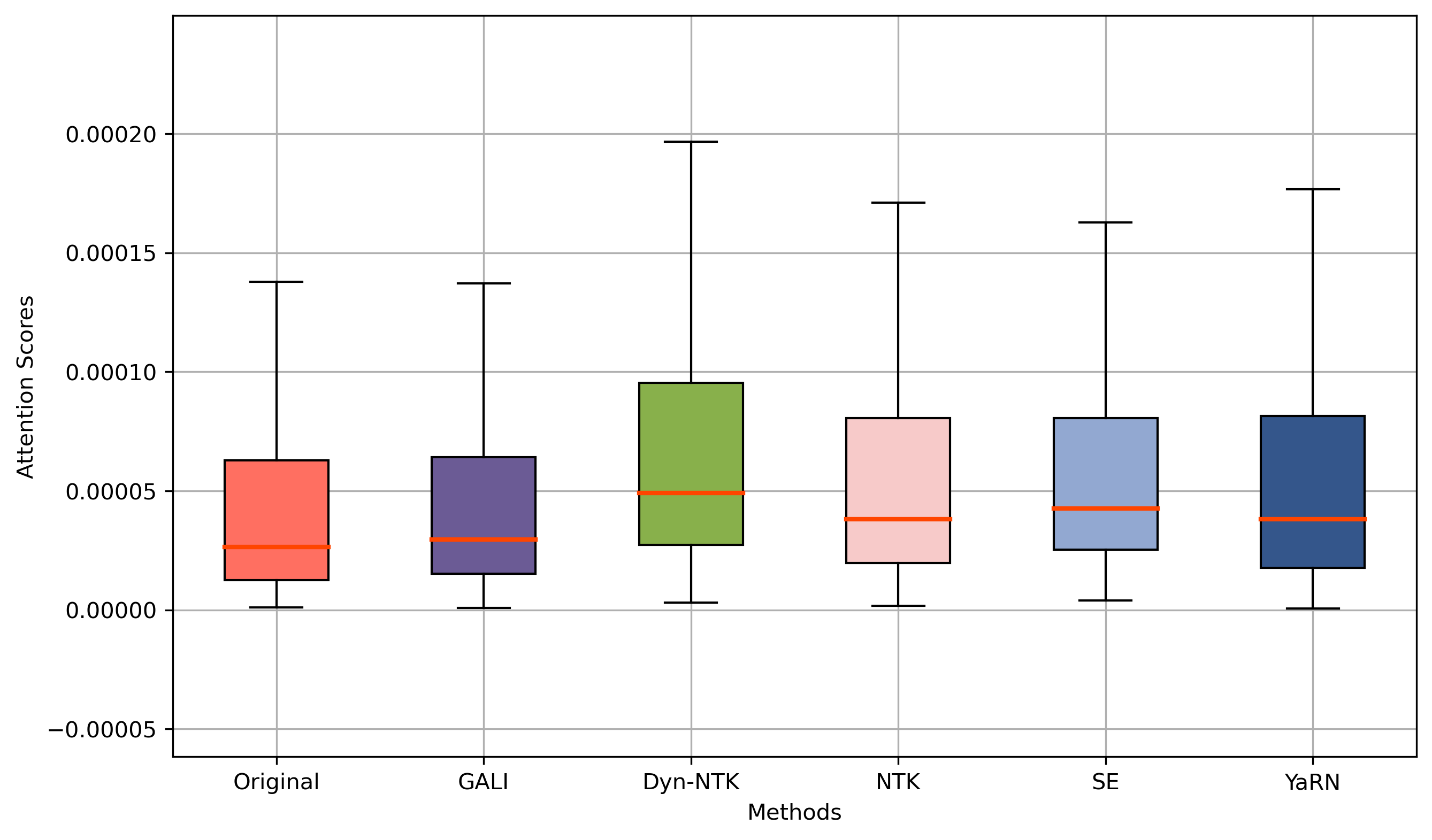}
        \label{fig:box_2k_3000}
    }
    \subfigure[Attention scores of row 4000]{
        \includegraphics[width=0.45\linewidth]{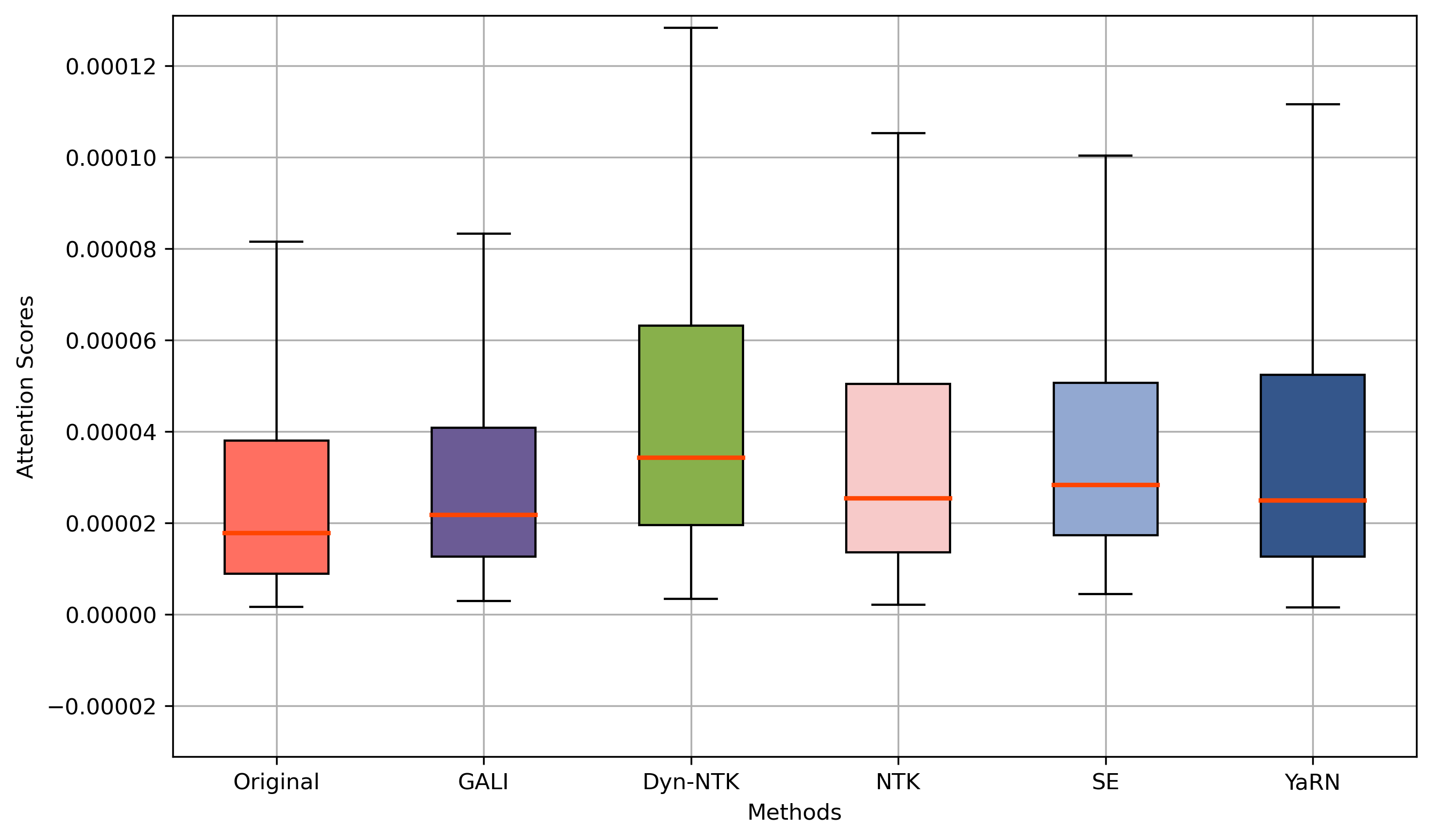}
        \label{fig:box_2k_4000}
    }
    \subfigure[Attention scores of row 5000]{
        \includegraphics[width=0.45\linewidth]{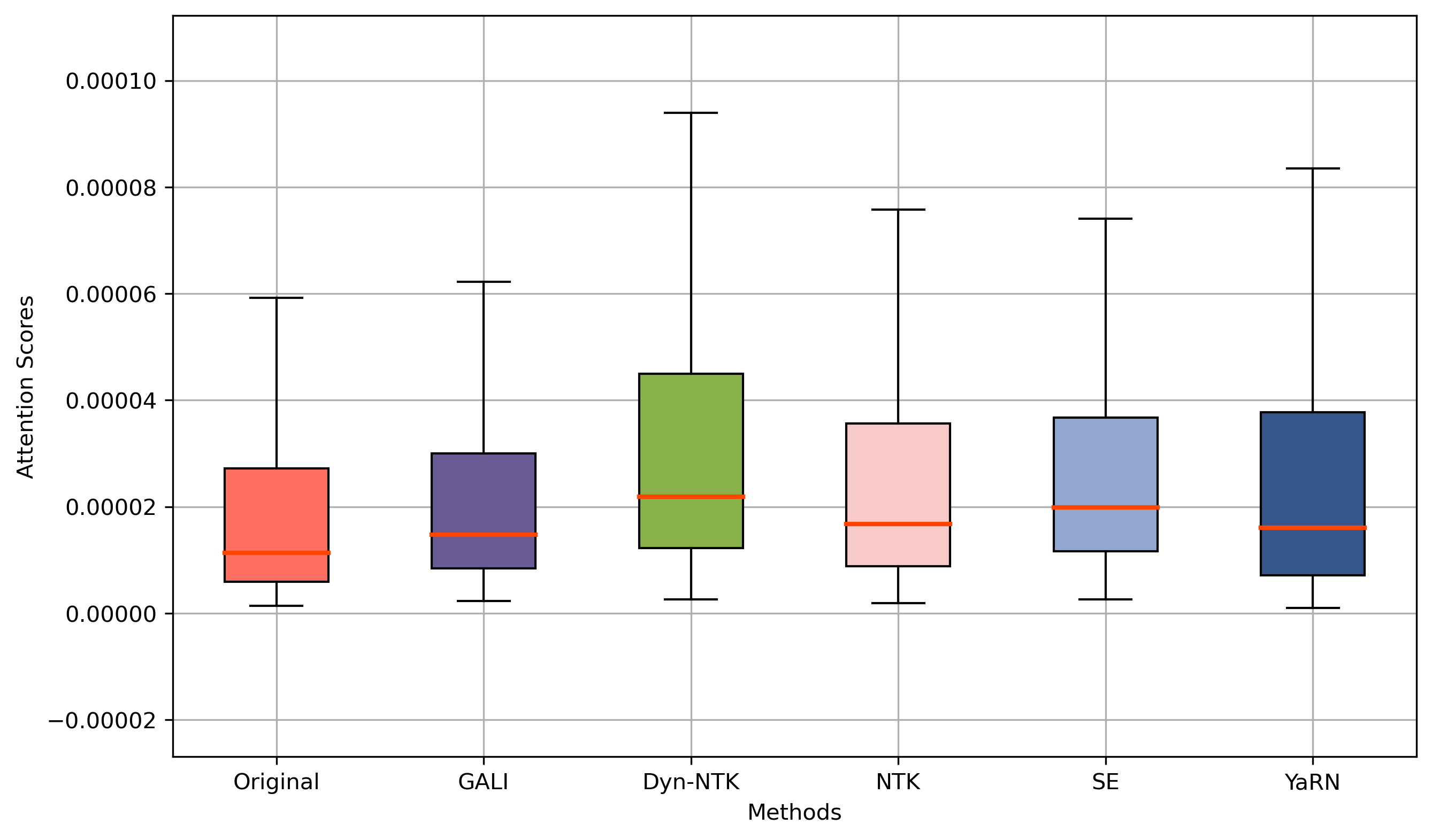}
        \label{fig:box_2k_5000}
    }
    \subfigure[Attention scores of row 6000]{
        \includegraphics[width=0.45\linewidth]{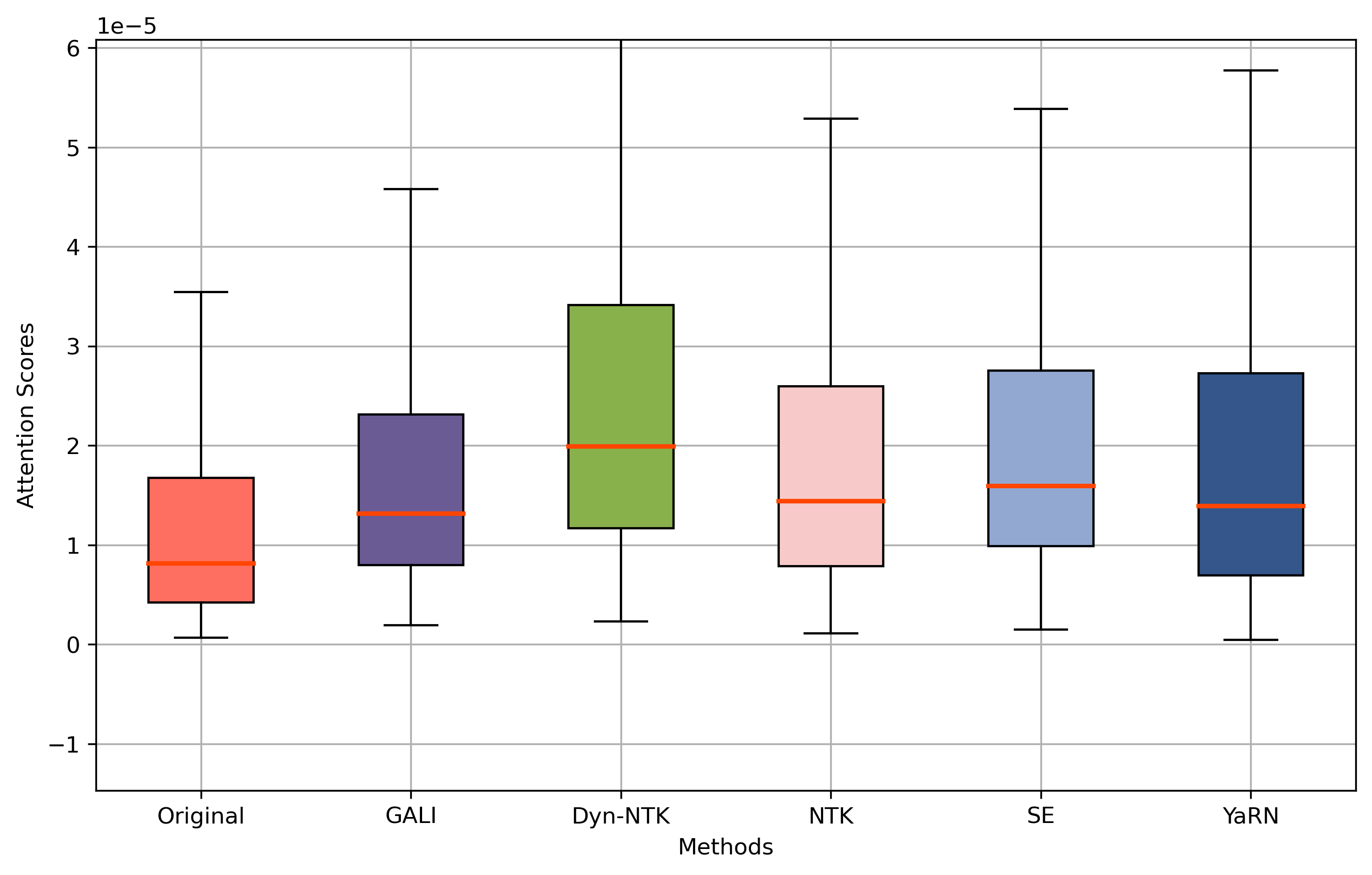}
        \label{fig:box_2k_6000}
    }
    \subfigure[Attention scores of row 7000]{
        \includegraphics[width=0.45\linewidth]{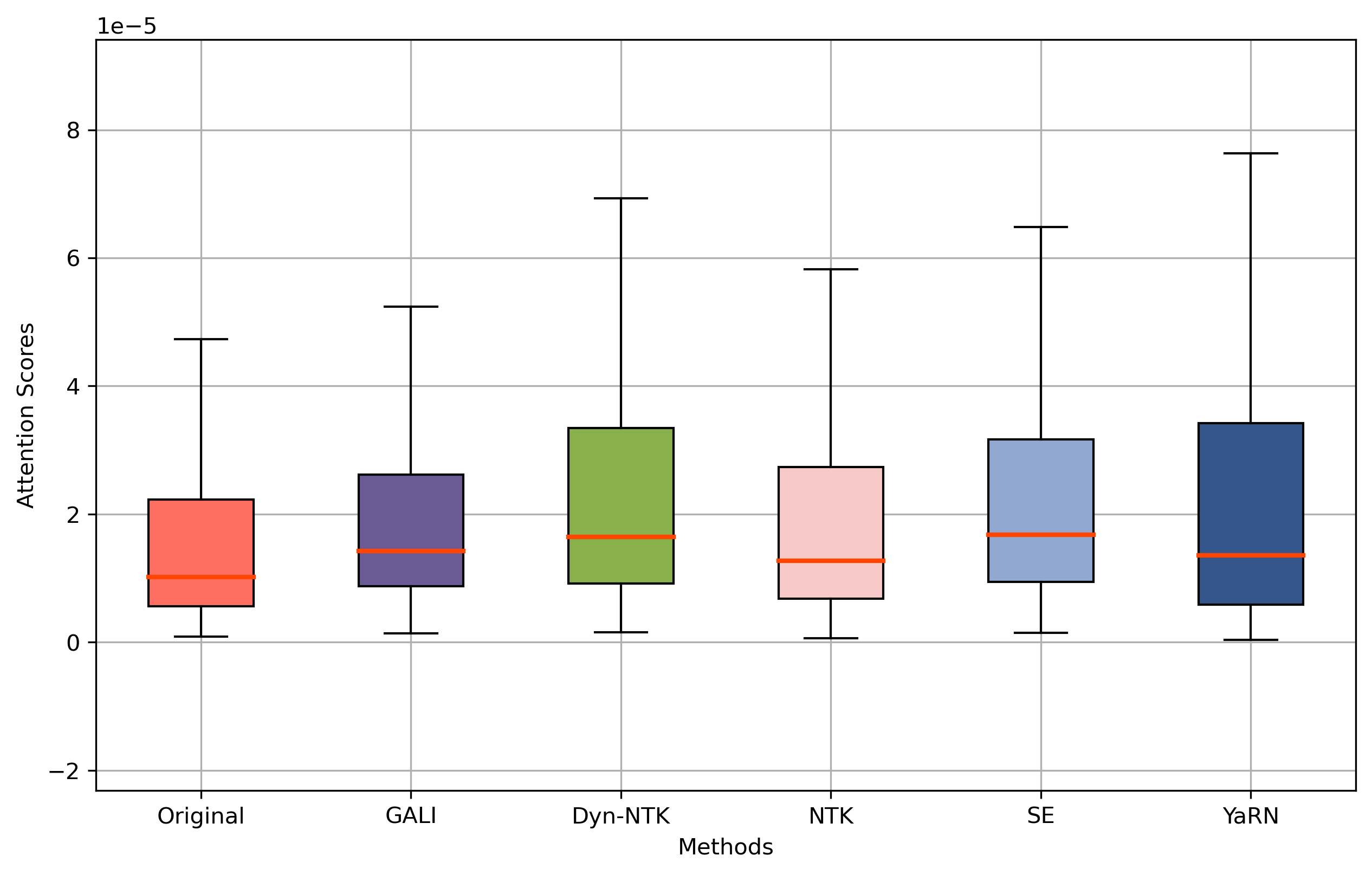}
        \label{fig:box_2k_7000}
    }
    \subfigure[Attention scores of row 8000]{
        \includegraphics[width=0.45\linewidth]{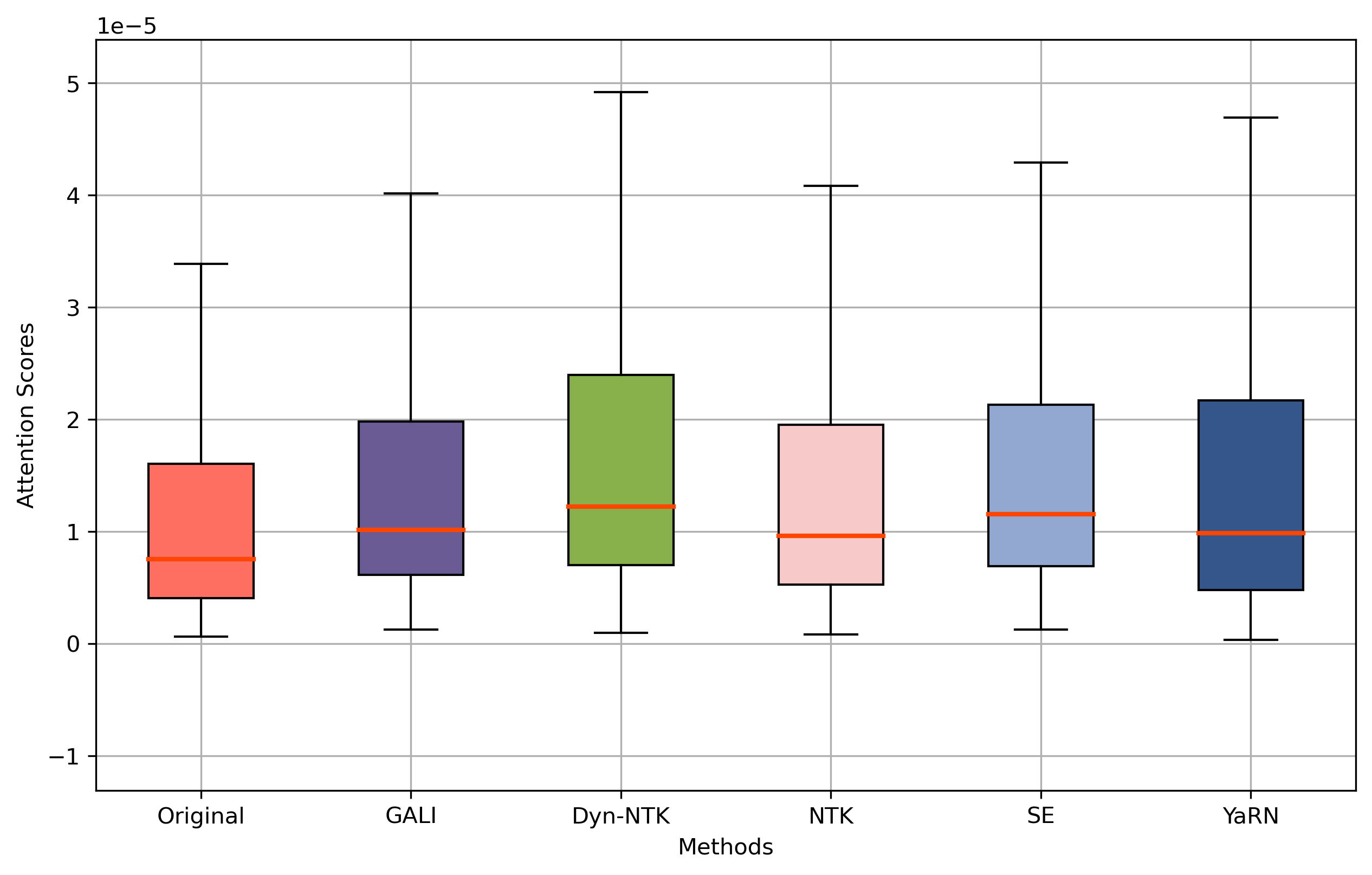}
        \label{fig:box_2k_8000}
    }
    \caption{Attention score distribution using Llama3-2k backbone. We omitted attention scores outside the 1st percentile and the 90th percentile here for clearer visualization.}
    \label{fig:2k_attn_rows}
\end{figure*}

\begin{figure*}[thp]
    \centering
    \subfigure[Attention scores of row 1000]{
        \includegraphics[width=0.45\linewidth]{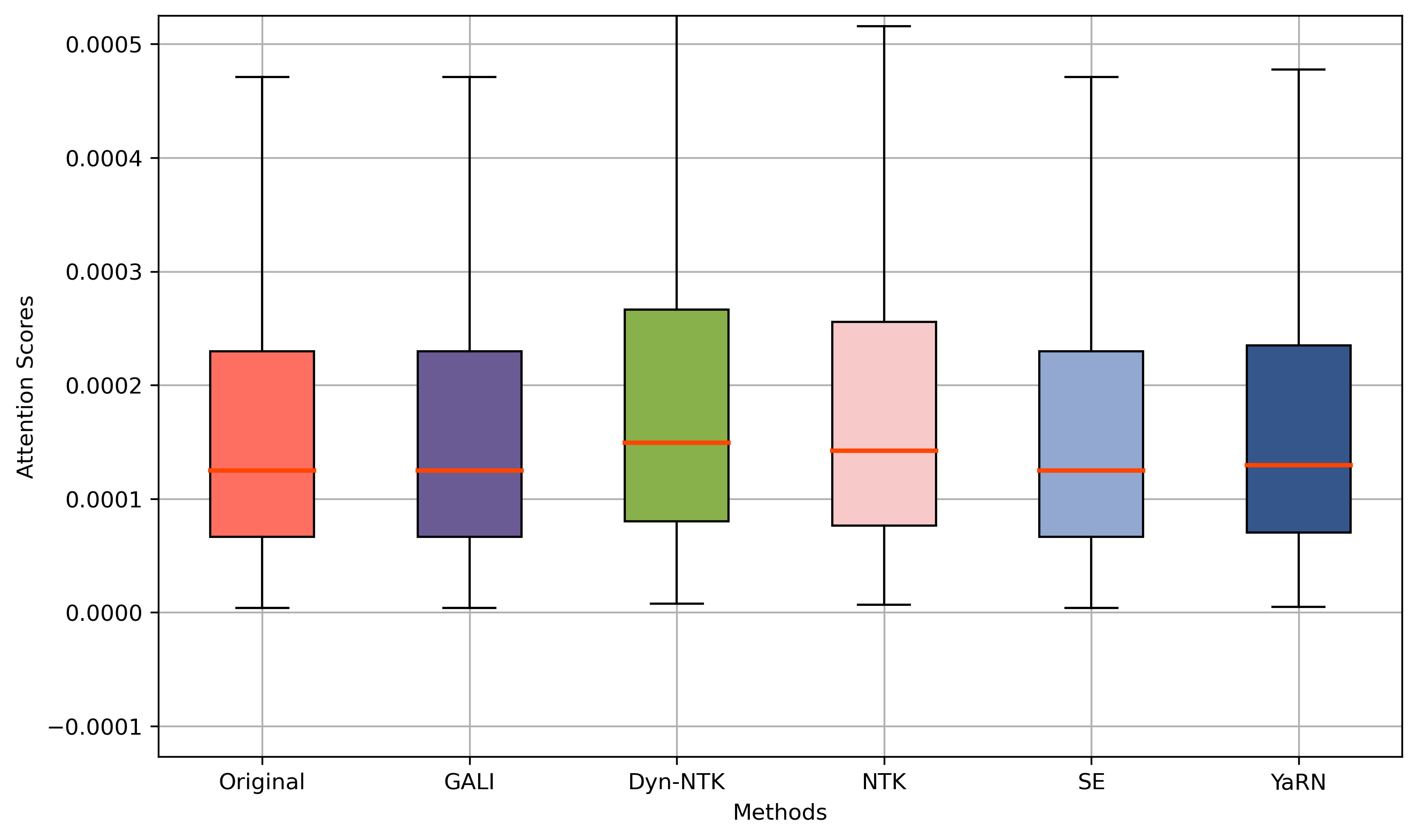}
        \label{fig:box_4k_1000}
    }
    \subfigure[Attention scores of row 2000]{
        \includegraphics[width=0.45\linewidth]{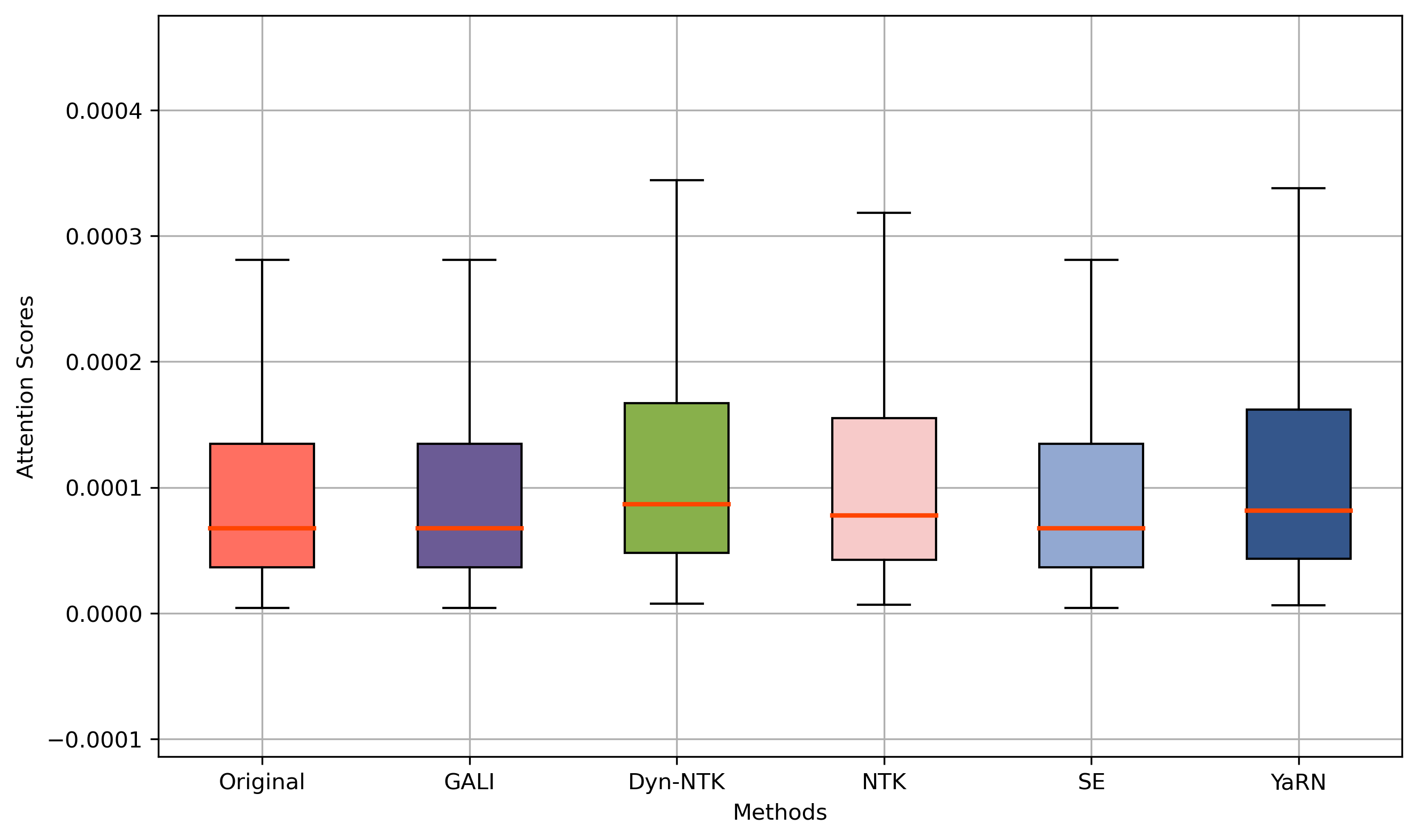}
        \label{fig:box_4k_2000}
    }
    \subfigure[Attention scores of row 3000]{
        \includegraphics[width=0.45\linewidth]{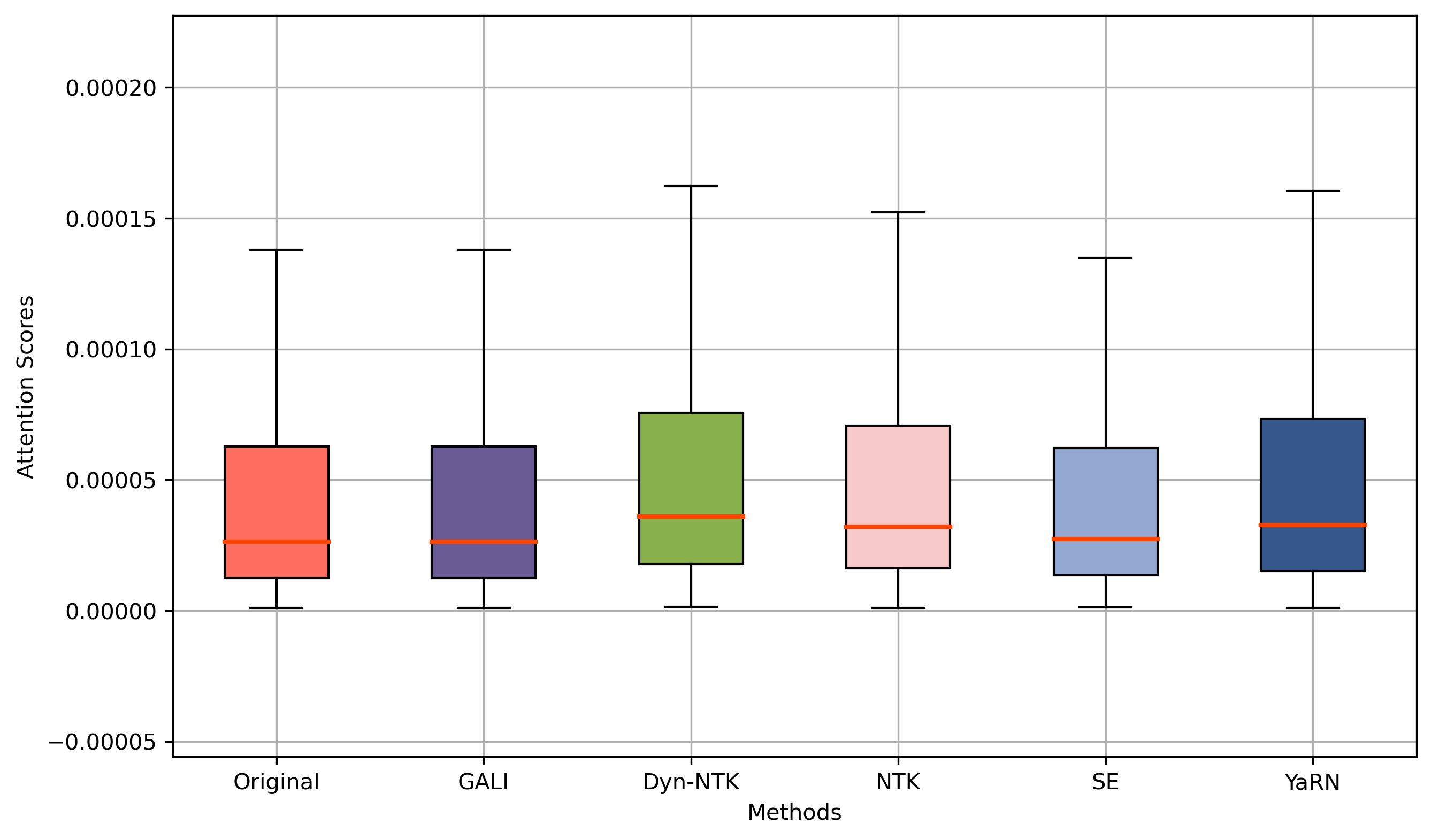}
        \label{fig:box_4k_3000}
    }
    \subfigure[Attention scores of row 4000]{
        \includegraphics[width=0.45\linewidth]{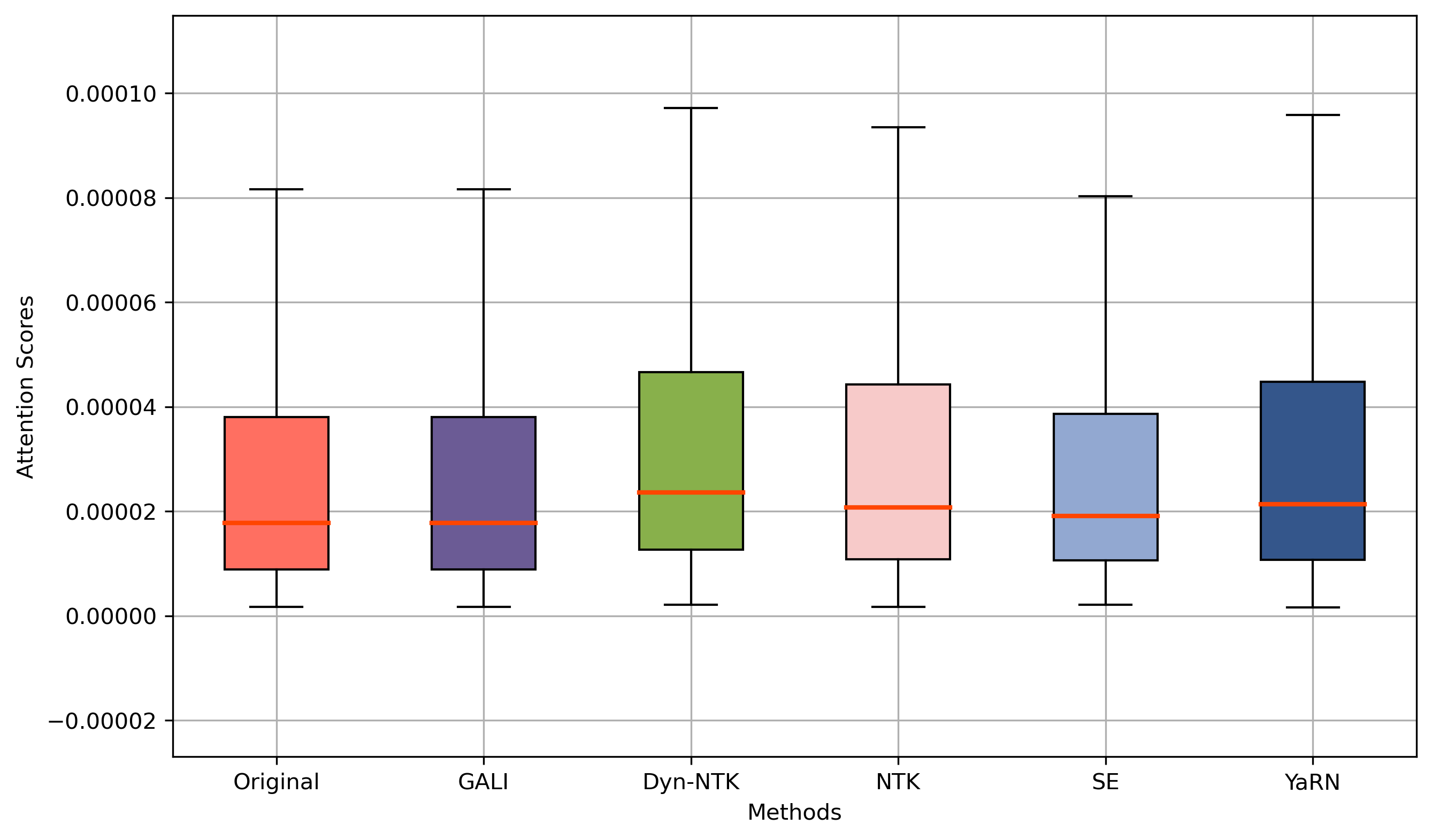}
        \label{fig:box_4k_4000}
    }
    \subfigure[Attention scores of row 5000]{
        \includegraphics[width=0.45\linewidth]{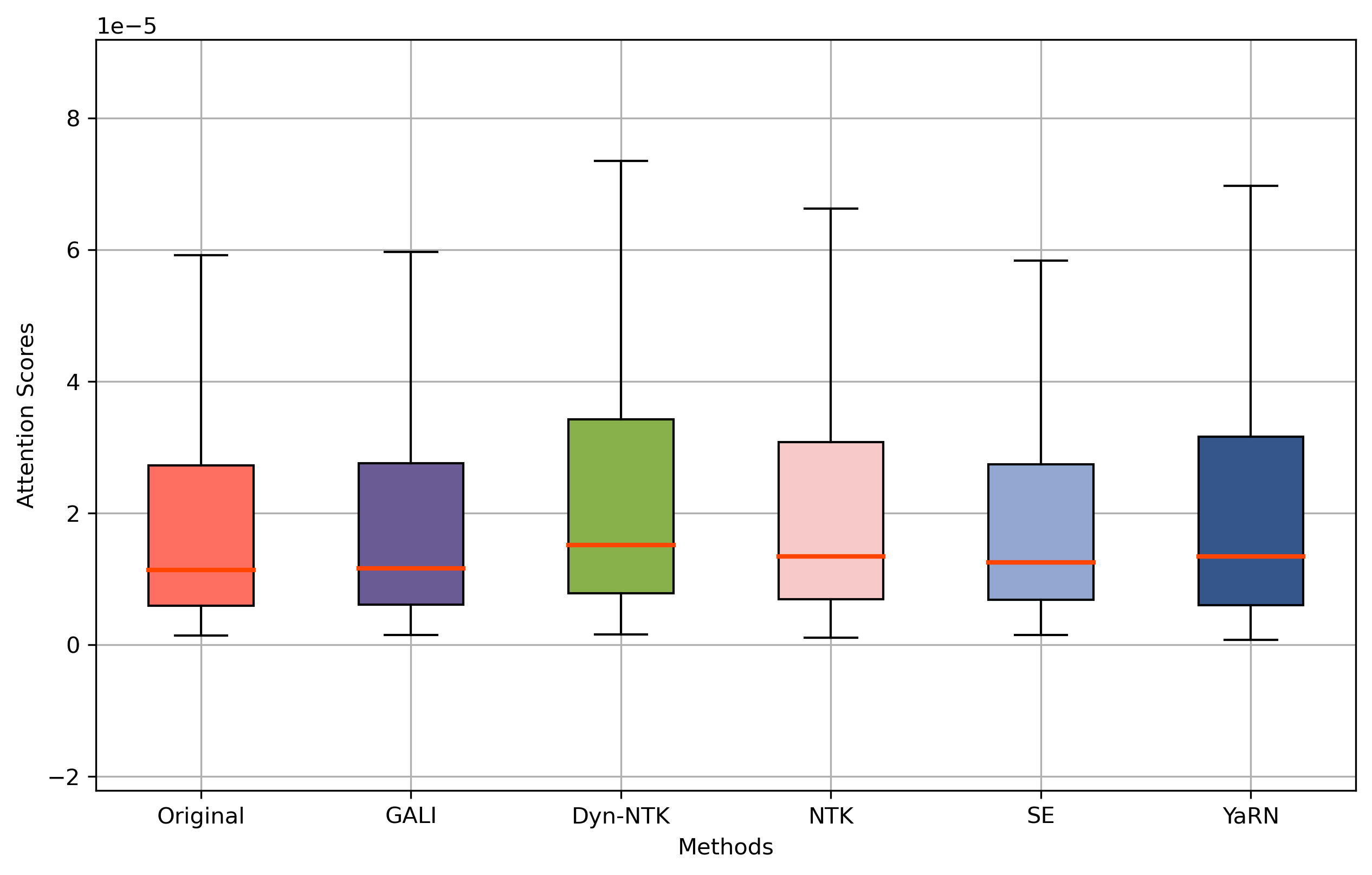}
        \label{fig:box_4k_5000}
    }
    \subfigure[Attention scores of row 6000]{
        \includegraphics[width=0.45\linewidth]{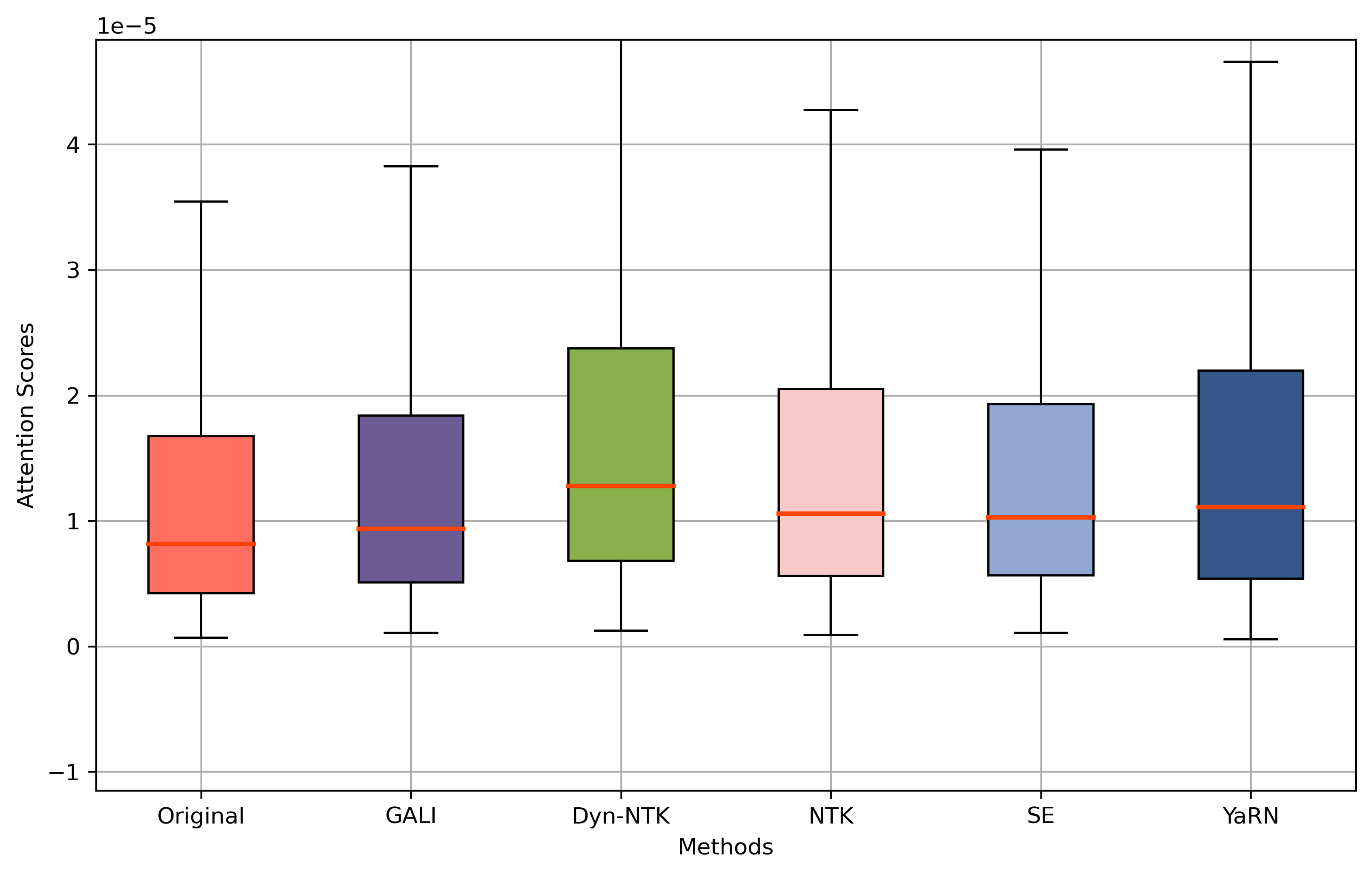}
        \label{fig:box_4k_6000}
    }
    \subfigure[Attention scores of row 7000]{
        \includegraphics[width=0.45\linewidth]{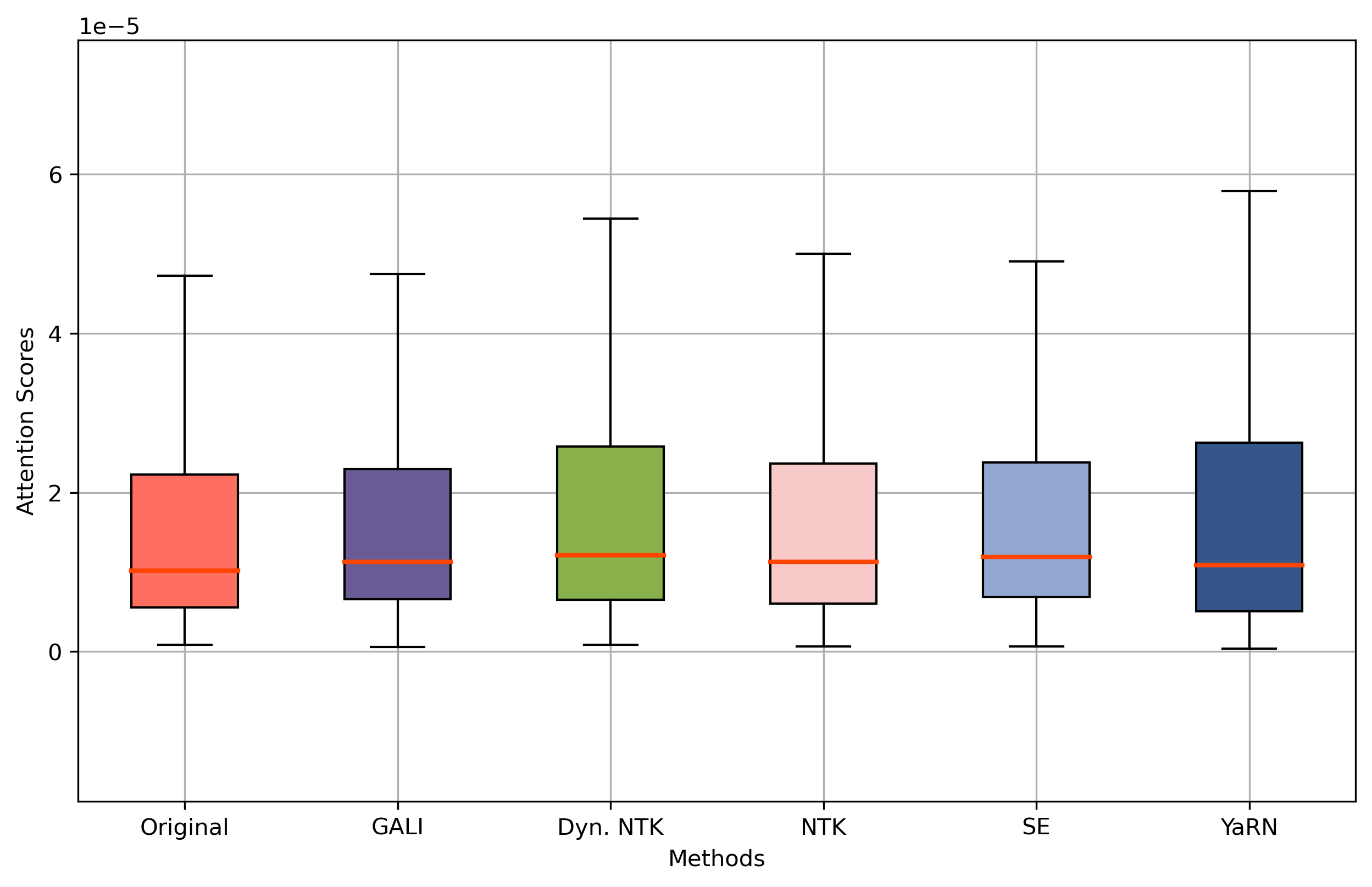}
        \label{fig:box_4k_7000}
    }
    \subfigure[Attention scores of row 8000]{
        \includegraphics[width=0.45\linewidth]{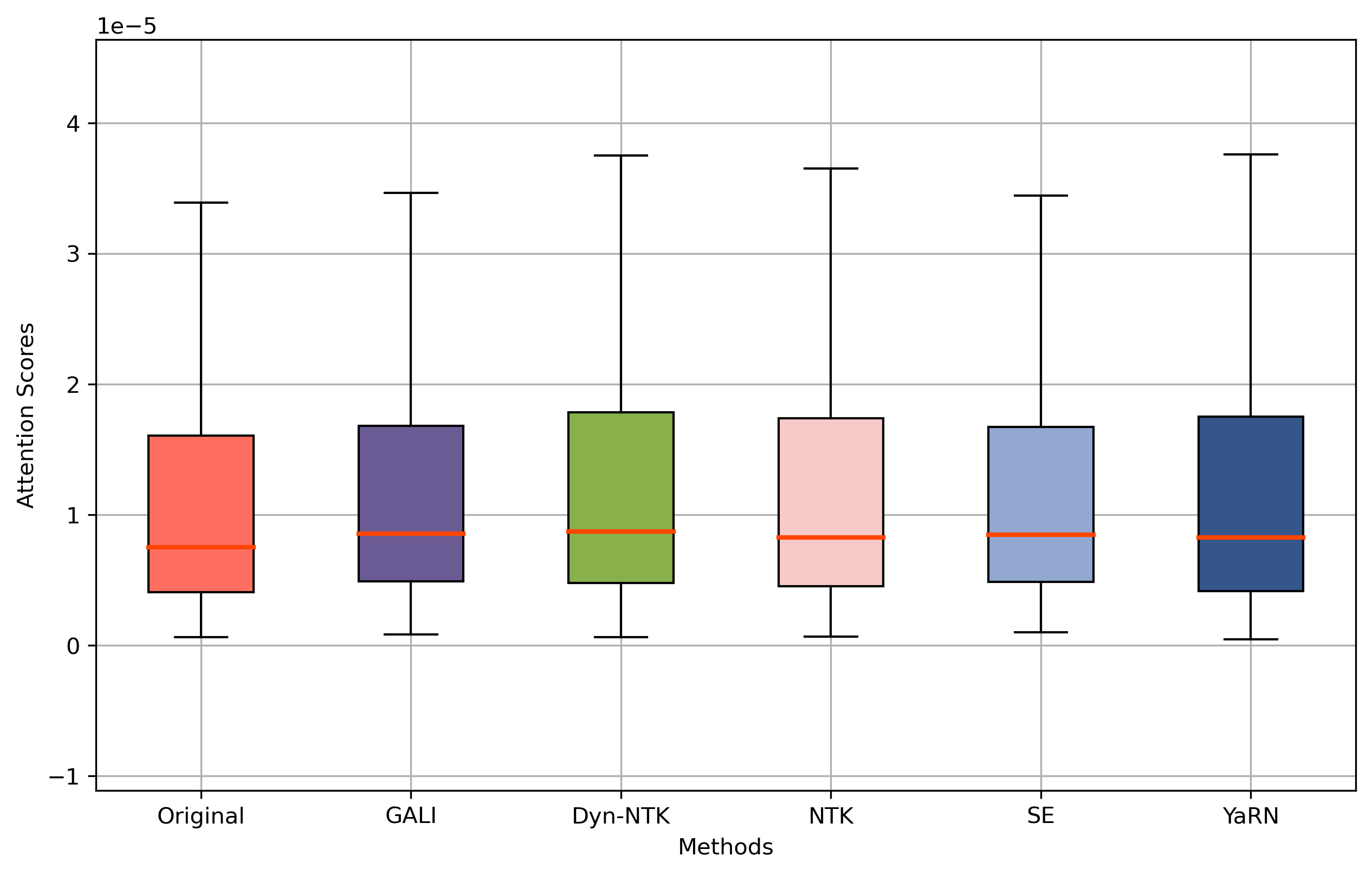}
        \label{fig:box_4k_8000}
    }
    \caption{Attention score distribution using Llama3-4k backbone. We omitted attention scores outside the 1st percentile and the 90th percentile here for clearer visualization.}
    \label{fig:4k_attn_rows}
\end{figure*}

\section{Pseudo code of GALI}
\label{app: gali pseudo code}
In this section, we provide the pseudo-code for the key steps required to implement GALI. Algorithm \ref{alg:get_chunk_size_list} generates the chunk sizes needed to partition the input during the prefill phase. While this function can be modified to support dynamic chunk sizes, we use fixed chunk sizes in our experiments to better control memory usage. Algorithm \ref{alg:construct_new_pi} interpolates new position IDs based on the minimum number of new IDs required for each chunk. Algorithm \ref{alg:attention_logit_interpolation} demonstrates how we perform attention logit interpolation. Note that we use $r = \lceil{m}\rceil - n$ to represent the interval between $q_m$ and $k_n$. This is because, when computing attention logit using RoPE, we cannot directly manipulate the relative positional interval matrix; instead, we modify the relative positional interval matrix by separately operating on $query\_states$ and $key\_states$. By using $r = \lceil{m}\rceil - n$, we ensure that $\lfloor{r}\rfloor = \lceil{m}\rceil - \lceil{n}\rceil$ and $\lceil{r}\rceil = \lceil{m}\rceil - \lfloor{n}\rfloor$, enabling modifications to the relative positional interval matrix while preserving the relative order between $query\_states$ and $key\_states$. It is important to note that some operations, such as reshaping, which do not affect the core concept, are omitted from the pseudo-code in these three algorithms.

\begin{algorithm*}[!htbp]
\centering
\caption{Generate Chunk Size List}
\label{alg:get_chunk_size_list}
\begin{algorithmic}[1]
\REQUIRE $prefill\_len$: The length of input in the prefill phase, $L_{tr}$: Training context window, $s$: Chunk size
\ENSURE A list of chunk sizes that sums to $prefill\_len$
\STATE $chunk\_size\_list \gets [L_{tr}]$
\STATE $sum\_len \gets L_{tr}$

\WHILE{$sum\_len < prefill\_len$}
    \STATE Append $s$ to $chunk\_size\_list$
    \STATE $sum\_len \gets sum\_len + s$
\ENDWHILE

\STATE Adjust the last chunk size:
\STATE $chunk\_size\_list[-1] \gets chunk\_size\_list[-1] - (sum\_len - prefill\_len)$

\STATE \textbf{return} $chunk\_size\_list$
\end{algorithmic}
\end{algorithm*}

\begin{algorithm*}[thp]
\caption{Position ID Interpolation}
\label{alg:construct_new_pi}
\begin{algorithmic}[1]
\REQUIRE $cur\_len$: Current length of the sequence, $L_{tr}$: Training context window, $add\_token$: The number of positions to be interpolated, $L_{w}$: Neighbor window size
\ENSURE $new\_pi$: New position IDs

\STATE $target\_len \gets cur\_len + add\_token$
\STATE $min\_group\_size \gets \lceil (target\_len - L_{w}) / (L_{tr} - L_{w}) \rceil$
\STATE $interval \gets 1 / min\_group\_size$
\STATE $total\_len \gets L_{tr}$
\STATE Initialize $new\_pi \gets []$ and $i \gets 0$

\WHILE{$total\_len < target\_len$}
    \STATE Append $[i + interval \cdot j \mid j \in \{0, 1, \dots, min\_group\_size - 1\}]$ to $new\_pi$ 
    \STATE $i \gets i + 1$ 
    \STATE $total\_len \gets L_{tr} - i + \text{len}(new\_pi)$
\ENDWHILE

\STATE $seg\_window \gets [j \mid j \in \{i, i+1, \dots, L_{tr} - 1\}]$
\STATE $new\_pi \gets new\_pi[:(target\_len - \text{len}(seg\_window))] + seg\_window$
\STATE \textbf{return} $new\_pi$
\end{algorithmic}
\end{algorithm*}

\begin{algorithm*}[!t]
\caption{Attention Logit Interpolation}
\label{alg:attention_logit_interpolation}
\begin{algorithmic}[1]
\REQUIRE $position\_ids$: Interpolated position IDs, $hidden\_states$: The inputs of the attention layer, $head\_dim$: The dimension of each head, $\text{q\_proj}$: Q project function, $\text{k\_proj}$: K project function, $\text{rotary\_emb}$: Rotary embedding function
\ENSURE Interpolated attention logit

\text{\# Compute the rotary embedding}
\STATE $cos\_ceil, sin\_ceil \gets \text{rotary\_emb}(hidden\_states, \lceil position\_ids \rceil)$
\STATE $cos\_floor, sin\_floor \gets \text{rotary\_emb}(hidden\_states, \lfloor position\_ids \rfloor)$

\text{\# Apply the rotary embedding on the query and key states}
\STATE $query\_states \gets \text{q\_proj}(hidden\_states)$
\STATE $key\_states \gets \text{k\_proj}(hidden\_states)$

\STATE $query\_states\_ceil \gets (query\_states \cdot cos\_ceil) + (\text{rotate\_half}(query\_states) \cdot sin\_ceil)$
\STATE $key\_states\_ceil \gets (key\_states \cdot cos\_ceil) + (\text{rotate\_half}(key\_states) \cdot sin\_ceil)$

\STATE $key\_states\_floor \gets (key\_states \cdot cos\_floor) + (\text{rotate\_half}(key\_states) \cdot sin\_floor)$

\text{\# Compute attention logit with $\lceil{R}\rceil and \lfloor{R}\rfloor$}
\STATE $attn\_floor \gets query\_states\_ceil \text{@} key\_states\_ceil^T / \sqrt{head\_dim}$
\STATE $attn\_ceil \gets query\_states\_ceil \text{@} key\_states\_floor^T / \sqrt{head\_dim}$

\STATE $rel\_coef \gets (\lceil position\_ids \rceil.unsqueeze(1) - position\_ids.unsqueeze(0)) \mod 1$

\STATE $attn\_logit \gets attn\_floor - (attn\_floor - attn\_ceil) \cdot rel\_coef$

\text{\# Add normal distribution noise}
\STATE $distance\_ids \gets [i \mid i \in \{0, 1, \dots, \text{len}(hidden\_states) - 1\}]$
\STATE $distance\_matrix \gets distance\_ids.unsqueeze(1) - distance\_ids.unsqueeze(0)$
\STATE $noise\_std \gets distance\_matrix / \text{len}(hidden\_states) $

\STATE $noise \gets \text{torch.randn\_like}(attn\_logit)$
\STATE $mask \gets (rel\_coef \neq 0)$

\STATE $noise \gets noise \cdot noise\_std \cdot mask$

\STATE $attn\_logit \gets attn\_logit + noise$

\STATE \textbf{return} $attn\_logit$
\end{algorithmic}
\end{algorithm*}

\section{Implementation details}
\label{app: imp. details}
In this section, we provide detailed implementation information for each method. For Dyn-NTK and YaRN, we utilize the implementations available in Huggingface\footnote{https://huggingface.co} by adding \textit{rope\_scaling = \{"rope\_type":"dynamic"\}} and \textit{rope\_scaling = \{"rope\_type":"yarn"\}}, respectively, to the LLM’s config.json file. For NTK, we implement it by adding \textit{rope\_scaling = \{"rope\_type":"dynamic"\}} and \textit{static\_ntk=True}, and modifying the \textit{\_dynamic\_frequency\_update} function of the LlamaRotaryEmbedding class as shown in Table \ref{tab:ntk_code}.


\begin{table*}[t]
\centering
\caption{The implementation of NTK used in our experiments.}
\label{tab:ntk_code}
\lstset{
  language=Python,           
  basicstyle=\ttfamily\footnotesize, 
  numberstyle=\tiny,         
  keywordstyle=\color{blue}, 
  commentstyle=\color{gray}, 
  stringstyle=\color{red},   
  frame=single,              
  breaklines=true,           
}
\begin{lstlisting}
def _dynamic_frequency_update(self, position_ids, device):
    """
    Modify this function to make it suitable for NTK
    """
    if self.config.static_ntk == True:
        if getattr(self, "reset_static_ntk", False) == False:
            config = copy.deepcopy(self.config)
            seq_len = self.original_max_seq_len * config.rope_scaling['factor']
            config.rope_scaling['factor'] = 1
            inv_freq, self.attention_scaling = self.rope_init_fn(
                config, device, seq_len=seq_len, **self.rope_kwargs
            )
            self.register_buffer("inv_freq", inv_freq, persistent=False) 
            setattr(self, "reset_static_ntk", True)
        return
    seq_len = torch.max(position_ids) + 1
    if seq_len > self.max_seq_len_cached:  # growth
        inv_freq, self.attention_scaling = self.rope_init_fn(
            self.config, device, seq_len=seq_len, **self.rope_kwargs
        )
        self.register_buffer("inv_freq", inv_freq, persistent=False) 
        self.max_seq_len_cached = seq_len

    if seq_len < self.original_max_seq_len and self.max_seq_len_cached > self.original_max_seq_len:  # reset
        self.register_buffer("inv_freq", self.original_inv_freq, persistent=False)
        self.max_seq_len_cached = self.original_max_seq_len

\end{lstlisting}
\end{table*}

For SelfExtend and ChunkLlama, we use their official implementations\footnote{SelfExtend: https://github.com/datamllab/LongLM, ChunkLlama: https://github.com/HKUNLP/ChunkLlama}. We list the hyperparameters required for these methods to extend to different maximum input length in Table \ref{tab:length_extrapolation_hyperparams}. All experiments can be conducted on a single A100 GPU (80GB) machine.

\begin{table*}[thp]
    \caption{Hyperparameters for length extrapolation methods under each setting. For example, “2k to 8k” indicates an initial context window of 2048, with a positional interval range of [0, 2048), and a target context window extending up to 8192. Other settings follow the same pattern. For GALI, the reported hyperparameters represent the combinations we search for each experiment.}
    \label{tab:length_extrapolation_hyperparams}
    \centering
    \begin{tabular}{|l|l|l|}
        \toprule
        \textbf{Exp.} & \textbf{Method} & \textbf{Hyperparameters} \\
        \midrule
        \multirow{6}{*}{2k to 8k} & NTK & rope\_scaling=\{”rope type”:”dynamic”, "factor": 4\}, static\_ntk=True \\
         & Dyn-NTK & rope\_scaling=\{”rope type”:”dynamic”, "factor": 4\} \\
         & YaRN & rope\_scaling=\{”rope type”:"YaRN", "factor": 4\} \\
         & SelfExtend & group\_size=5, window\_size=512 \\
         & ChunkLlama & chunk\_size=1536, local\_window=128 \\
         & GALI & chunk\_size=[1000,2000,3000], local\_window=[128, 256, 512, 1024] \\
        \midrule
        \multirow{6}{*}{4k to 8k} & NTK & rope\_scaling=\{”rope type”:”dynamic”, "factor": 2\}, static\_ntk=True \\
         & Dyn-NTK & rope\_scaling=\{”rope type”:”dynamic”, "factor": 2\} \\
         & YaRN & rope\_scaling=\{”rope type”:"YaRN", "factor": 2\} \\
         & SelfExtend & group\_size=3, window\_size=2048 \\
         & ChunkLlama & chunk\_size=3072, local\_window=256 \\
         & GALI & chunk\_size=[1000,2000,3000], local\_window=[128, 256, 512, 1024] \\
        \midrule
        \multirow{6}{*}{4k to 16k} & NTK & rope\_scaling=\{”rope type”:”dynamic”, "factor": 4\}, static\_ntk=True \\
         & Dyn-NTK & rope\_scaling=\{”rope type”:”dynamic”, "factor": 4\} \\
         & YaRN & rope\_scaling=\{”rope type”:"YaRN", "factor": 4\} \\
         & SelfExtend & group\_size=5, window\_size=1024 \\
         & ChunkLlama & chunk\_size=3072, local\_window=256 \\
          & GALI & chunk\_size=[1000,2000,3000], local\_window=[128, 256, 512, 1024] \\
        \midrule
        \multirow{6}{*}{4k to 32k} & NTK & rope\_scaling=\{”rope type”:”dynamic”, "factor": 8\}, static\_ntk=True \\
         & Dyn-NTK & rope\_scaling=\{”rope type”:”dynamic”, "factor": 8\} \\
         & YaRN & rope\_scaling=\{”rope type”:"YaRN", "factor": 8\} \\
         & SelfExtend & group\_size=15, window\_size=2048 \\
         & ChunkLlama & chunk\_size=3072, local\_window=256 \\
        & GALI & chunk\_size=[1000,2000,3000], local\_window=[128, 256, 512, 1024] \\
        \midrule
        \multirow{6}{*}{8k to 16k} & NTK & rope\_scaling=\{”rope type”:”dynamic”, "factor": 2\}, static\_ntk=True \\
         & Dyn-NTK & rope\_scaling=\{”rope type”:”dynamic”, "factor": 2\} \\
         & YaRN & rope\_scaling=\{”rope type”:"YaRN", "factor": 2\} \\
         & SelfExtend & group\_size=3, window\_size=4096 \\
         & ChunkLlama & chunk\_size=6144, local\_window=512 \\
        & GALI & chunk\_size=[1000,2000,3000], local\_window=[128, 256, 512, 1024] \\
        \midrule
        \multirow{6}{*}{8k to 32k} & NTK & rope\_scaling=\{”rope type”:”dynamic”, "factor": 4\}, static\_ntk=True \\
         & Dyn-NTK & rope\_scaling=\{”rope type”:”dynamic”, "factor": 4\} \\
         & YaRN & rope\_scaling=\{”rope type”:"YaRN", "factor": 4\} \\
         & SelfExtend & group\_size=5, window\_size=2048 \\
         & ChunkLlama & chunk\_size=6144, local\_window=512 \\
        & GALI & chunk\_size=[1000,2000,3000], local\_window=[128, 256, 512, 1024] \\
        \bottomrule
    \end{tabular}
\end{table*}

\end{document}